\documentclass[lettersize,journal]{IEEEtran}
\usepackage{amsmath,amsfonts}
\usepackage{algorithmic}
\usepackage{algorithm}
\usepackage{array}
\usepackage[caption=false,font=normalsize,labelfont=sf,textfont=sf]{subfig}
\usepackage{textcomp}
\usepackage{stfloats}
\usepackage{url}
\usepackage{verbatim}
\usepackage{graphicx}
\usepackage{cite}
\usepackage[colorlinks=true]{hyperref}
\hypersetup{
  linkcolor=black,
  citecolor=black,
  urlcolor=[RGB]{5,115,185}
}
\usepackage{mathrsfs}
\usepackage{mathtools}
\usepackage{amsthm}
\usepackage{stmaryrd}
\usepackage{tabularx}
\usepackage{booktabs}
\usepackage{orcidlink}

\usepackage{tabularx}
\usepackage{booktabs}
\usepackage{array}
\usepackage{pifont}
\newcolumntype{L}{>{\raggedright\arraybackslash}X}   
\newcolumntype{Y}{>{\centering\arraybackslash}X}

\usepackage{xcolor}    
\usepackage[justification=centering]{caption}

\usepackage{fancyhdr}

\fancypagestyle{IEEEtitlepagestyle}{%
  \fancyhf{}%
  \fancyfoot[C]{\thepage} 
}

\newcolumntype{a}{>{\columncolor{Green}}c}
\newcolumntype{b}{>{\columncolor{DarkOrange}}c}

\newtheorem{theorem}{Theorem}[section]
\newtheorem{remark}{Remark}

\newtheorem{definition}[theorem]{Definition}

\usepackage{etoolbox}
\makeatletter
\patchcmd{\@begintheorem}{\textit}{\textbf}{}{}
\makeatother

\renewcommand\thesection{\arabic{section}}
\def\thesectiondis{\thesection.}                   
\def\thesubsectiondis{\thesectiondis\arabic{subsection}.}

\renewcommand{\thetable}{\arabic{table}}

\numberwithin{equation}{section}
\numberwithin{theorem}{section}
\numberwithin{proposition}{section}
\numberwithin{remark}{section}
\numberwithin{problem}{section}
\numberwithin{subsection}{section}

\usepackage{etoolbox}

\AtBeginEnvironment{tabular}{\small}

\def\loss{\mathrm{loss}}

\newcommand{\inter}[1]{\llbracket #1 \rrbracket}

\def\*#1{\boldsymbol{#1}}
\makeatletter
\newcommand{\vx}{\vec{x}\@ifnextchar{^}{\,}{}}
\makeatother

\begin{document}

\title{The \href{https://sherpa.ai/}{Sherpa.ai} Blind Vertical Federated Learning Paradigm to Minimize the Number of Communications}




\author{{\vspace{2.7em} 
{\LARGE \href{https://sherpa.ai/}{Sherpa.ai}}\\[1ex]
research@sherpa.ai}}



\maketitle

\begin{abstract}
Federated Learning (FL) enables collaborative decentralized training across multiple parties (nodes) while keeping raw data private. There are two main paradigms in FL: Horizontal FL (HFL), where all participant nodes share the same feature space but hold different samples, and Vertical FL (VFL), where participants hold complementary features for the same samples. While HFL is widely adopted, VFL is employed in domains where nodes hold complementary features about the same samples. Still, VFL presents a significant limitation: the vast number of communications required during training. This compromises privacy and security, and can lead to high energy consumption, and in some cases, make model training unfeasible due to the high number of communications.

In this paper, we introduce Sherpa.ai Blind Vertical Federated Learning (SBVFL), a novel paradigm that leverages a distributed training mechanism enhanced for privacy and security. Decoupling the vast majority of node updates from the server dramatically reduces node-server communication. Experiments show that SBVFL reduces communication by $\sim$99\% compared to standard VFL while maintaining accuracy and robustness. Therefore, SBVFL enables practical, privacy-preserving VFL across sensitive domains, including healthcare, finance, manufacturing, aerospace, cybersecurity, and the defense industry.
\end{abstract}


\section{Introduction}\label{sec:intro}

Federated Learning (FL)~\cite{mcmahan2017communication} enables collaborative training across multiple nodes (parties, clients, devices) while keeping raw data decentralized, sharing only model updates instead of centralizing data as in traditional Machine Learning (ML). FL is typically categorized into Horizontal FL (HFL), where nodes share the same feature space but hold different samples, and Vertical FL (VFL), where nodes hold data with different feature spaces for the same set of samples~\cite{wen2023survey}. 

Sherpa.ai Blind VFL (SBVFL) is a novel VFL paradigm developed within the research team at Sherpa.ai to address the challenges of healthcare, security, and telecommunication systems. It has been specifically designed to achieve ad hoc performance in VFL by minimizing the communication between nodes and servers.

The cornerstone of this novel approach is the possibility for the nodes to perform independent training without the need for repeated exchanges of information about gradients and parameter updates with the central server. This is made possible by providing each node with \textit{synthetic (surrogate) labels}, which are generated on the server via a secret procedure that prevents clients from inferring the true training labels; the server later maps them back to the original labels to complete training. In this way, SBVFL reduces the number of communication rounds\footnote{\textbf{Definition.} Let $H$ be the set of training variables needed to carry out a federated training. The minimal set of communications needed to transmit all variables in $H$ is referred to as a communication round.
} while achieving predictive performance comparable to that of centralized training, and drastically enhancing privacy and security.
	
Our simulation experiments, conducted using a Residual Neural Network (ResNet) framework, show that SBVFL can complete a classification task with high accuracy at low computational cost. Furthermore, SBVFL is a general approach, applicable to a wide spectrum of VFL systems in which the nodes may have diverse model architectures.

\subsection{Motivation and Challenges}
A major obstacle in VFL is the vast number of communication rounds required between the nodes and the server during training~\cite{li2023fedvs,ren2022improving,chen2020vafl}. These repeated exchanges not only increase energy consumption and system latency but also exacerbate the risks of privacy leakage through gradient or label-inference attacks~\cite{fu2022label}. This high communication overhead is especially problematic in scenarios with strict bandwidth limitations (e.g., satellite or IoT networks) or where energy efficiency is critical (e.g., mobile or defense applications)~\cite{qiu2020planet}. 

Recent efforts have proposed cryptographic approaches, such as Homomorphic Encryption (HE) or Secure Multi-Party (SMPC) Computation, to mitigate privacy risks~\cite{hardy2017private,gascon2016secure}. However, these methods introduce substantial computational costs and scalability issues. Hence, there is a pressing need for VFL frameworks that simultaneously minimize communication, preserve privacy, and maintain competitive predictive performance.

\subsection{Contributions}
We propose the SBVFL paradigm, a novel framework designed to address the inefficiencies of traditional VFL. Our main contributions are summarized as follows:
\begin{itemize}
    \item We introduce a blind learning paradigm that utilizes server-generated synthetic (surrogate) labels, allowing nodes to train independently while significantly reducing the number of required communication rounds.
    \item We provide a rigorous theoretical analysis, demonstrating that SBVFL retains the classification capability of centralized models while improving privacy guarantees.
    \item We evaluate SBVFL on real-world datasets, including image classification and financial default prediction, demonstrating that our approach drastically reduces communication by $\sim$99\% compared to standard VFL while achieving competitive accuracy.
    \item We highlight SBVFL’s practical advantages for high-stakes applications such as healthcare, finance, and national defense, where data confidentiality, communication efficiency, and robustness against adversarial attacks are paramount.
\end{itemize}

The paper is organized as follows: Section~\ref{sec:related_work} shows the related work. Section~\ref{sec:preliminaries} reviews key concepts in FL, focusing on VFL and establishing notation. Section~\ref{sec:fake_labels} introduces synthetic label generation as a preliminary step for SBVFL, detailed in Section~\ref{sec:BVFL}. Section~\ref{sec:control} provides a theoretical analysis using simultaneous controllability for neural ordinary differential equations (Neural ODEs), while Section~\ref{sec:privacy} discusses SBVFL’s privacy-preserving properties. Section~\ref{sec:experimental_results} presents experimental results, and Section~\ref{sec:Real world use cases and benefits} highlights a real-world application. Section~\ref{sec:Limitations} discusses limitations, and Section~\ref{sec:Conclusion} concludes with future work.

\section{Related work}
\label{sec:related_work}

We provide an overview of related work in VFL, with emphasis on communication efficiency, privacy, and security, while situating our contribution within the broader literature.  

\subsection{Communication Efficiency in VFL}
A major bottleneck in VFL is the large number of communication rounds required to exchange intermediate representations (embedding outputs) and gradients between nodes and the server. Several works have attempted to reduce synchronization costs. Asynchronous and straggler-resilient protocols, such as VAFL~\cite{chen2020vafl} and FedVS~\cite{li2023fedvs}, mitigate delays by relaxing synchronous updates; however, they still require frequent interactions tied to each mini-batch, leading to high overhead. Split- or teacher–student-style variants introduce soft targets to amortize server interactions~\cite{ren2022improving}, although these signals can inadvertently leak label information.  

Beyond scheduling, techniques from HFL have inspired communication-reduction strategies that can be transferred to VFL, such as gradient compression, quantization, and sparsification~\cite{chen2018lag, mishchenko2022proxskip}. While these methods reduce traffic, they typically maintain per-batch coordination, which limits scalability in scenarios where bandwidth or latency is constrained. In contrast, our approach decouples most node training from server feedback, resulting in significantly fewer communication rounds.

\subsection{Privacy and Security in VFL}
Even when raw data remain local, exchanged activations or gradients may leak sensitive information. Label inference attacks~\cite{fu2022label} have demonstrated that server-owned labels can sometimes be reconstructed from node–server communications. To mitigate such risks, cryptographic safeguards like HE and SMPC have been applied to VFL~\cite{hardy2017private, gascon2016secure}. However, these approaches often introduce high computational and communication costs, limiting their practicality in large-scale deployments. Hybrid methods that integrate differential privacy (DP)~\cite{wang2020hybrid} or secure entity resolution~\cite{xu2021fedv} offer stronger confidentiality but remain costly under real-world constraints.  

Recent work seeks to limit label exposure entirely. BlindFL~\cite{fu2022blindfl} and related methods~\cite{salmeron2024blind, razavikia2024blind} differ from our use of “blind”: Fu et al.~\cite{fu2022blindfl} depend on HE/Secret Sharing, inflating communication; Salmeron et al.~\cite{salmeron2024blind} describe an HFL scheme without an initial model; and Razavikia et al.~\cite{razavikia2024blind} target an over-the-air HFL setting. By contrast, SBVFL utilizes server-generated synthetic labels, preserving label privacy without requiring heavy cryptography and eliminating the need for continuous server supervision, thereby reducing communication by orders of magnitude while maintaining competitive accuracy.

\subsection{Positioning and Applications}
Within this landscape, recent surveys~\cite{liu2024vertical, khan2025vertical} highlight the key challenge of jointly optimizing utility, privacy, and communication efficiency in VFL. Existing methods either reduce communication while risking label leakage or strengthen privacy at the expense of scalability. SBVFL advances the state of the art by achieving both high privacy and drastic communication reduction ($\sim$99\%) without compromising accuracy.  

These advances matter most where confidentiality and bandwidth are tight constraints: in healthcare, FL shows promise across hospitals but suffers from communication overhead~\cite{chen2020fedhealth}; in finance, adoption for credit risk and fraud is curtailed by regulation and cross-organizational sharing costs~\cite{yang2019ffd}; and in telecom/IoT, low bandwidth and energy limits dominate~\cite{qiu2020planet}. SBVFL addresses these issues with a communication-light, privacy-preserving approach that is well-suited to such settings.  

In summary, prior works have significantly advanced VFL in communication scheduling, cryptographic safeguards, and privacy-preserving protocols. However, none simultaneously minimize communication, preserve privacy, and maintain predictive performance at scale. Our proposed SBVFL framework aims to bridge this gap.

\section{Background and Problem Formulation}\label{sec:preliminaries}

FL \cite{mcmahan2017communication,chen2020fedhealth} is an emerging framework that aims to train an ML model (e.g., a deep neural network) on multiple local datasets located in different nodes, without explicitly exchanging data samples. This approach stands in contrast to traditional \textit{centralized} ML techniques, where all local datasets are aggregated and uploaded to a single server. FL, instead, enables multiple nodes to build a common learning model under the supervision of a central \textit{server}, while respecting privacy protocols like the EU General Data Protection Regulation \cite{GDPR}, the U.S. National AI Initiative \cite{UAI}, or the UK Data Protection Act \cite{KAI}. Its applications span a large number of fields, including defense, telecommunications, personalized healthcare, and IoT \cite{yang2019federated}.
	
\begin{figure}[ht]
        \centering
	\includegraphics[width=\columnwidth]{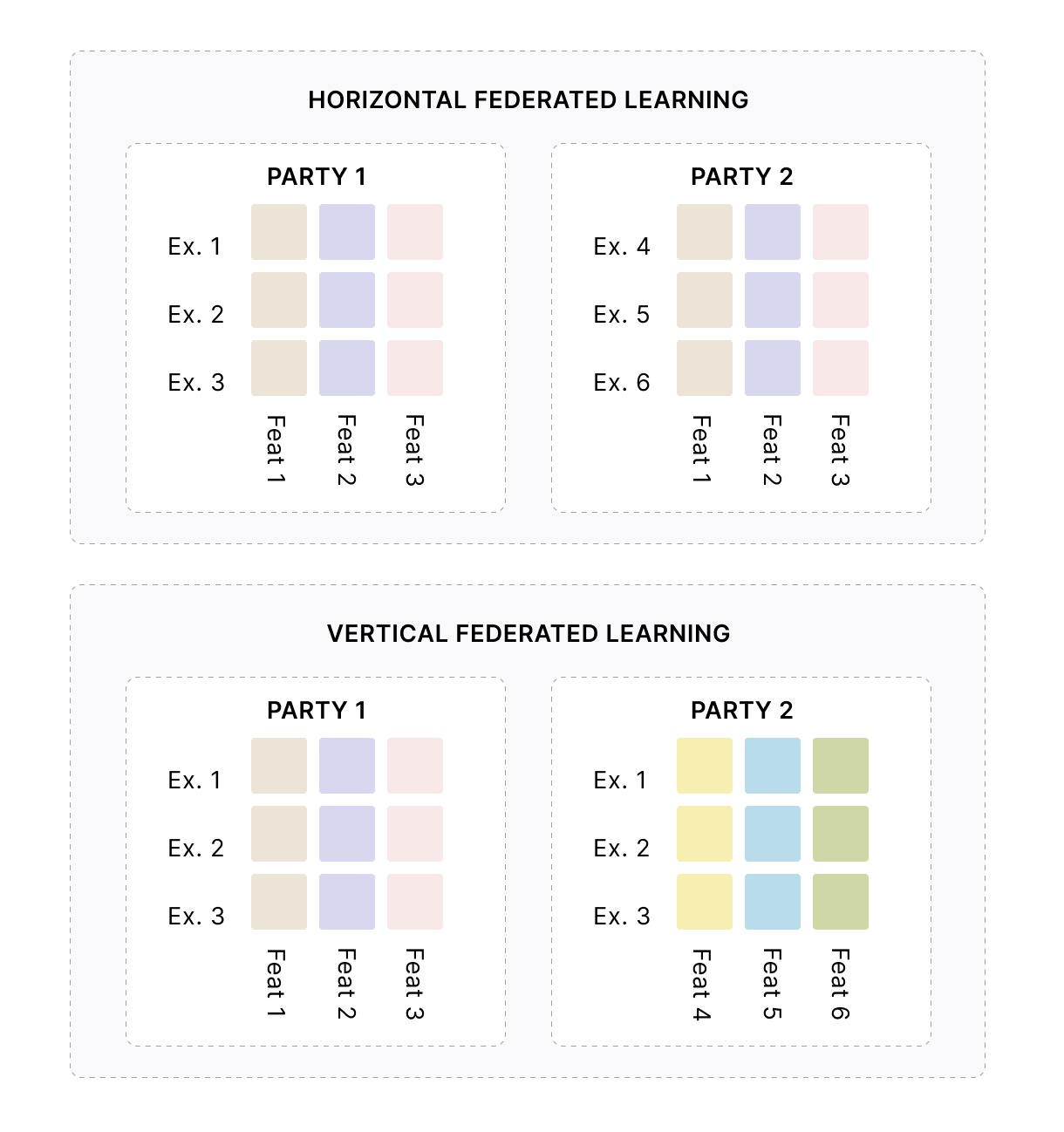}
	\caption{Graphical description of the HFL and VFL scenarios. 
    }
        \label{fig:VFL_scheme}
\end{figure}

	FL is typically subdivided into two broad categories, depending on the data distribution (see Figure \ref{fig:VFL_scheme}):
\begin{itemize}
    \item \textbf{HFL}~\cite{mcmahan2017communication,wang2021field,mishchenko2022proxskip}, also known as the homogeneous data scenario, in which different nodes have the same feature space but little intersection in the sample space. This is a natural data partitioning when different users collaboratively train a model on the same task using FL. A typical example is a federation of hospitals collaboratively training a cancer detection model using chest X-ray images of different patients. Here, the sample IDs correspond to the individual patients in each hospital (with little or no overlap across institutions), while the feature space is the pixel intensity values of the X-ray images, which are common across all hospitals.
    \item \textbf{VFL}~\cite{gascon2016secure,hardy2017private,wang2020hybrid,chen2020vafl,xu2021fedv,ren2022improving,li2023fedvs,liu2024vertical,folino2024efficiently,khan2025vertical}, also known as the heterogeneous data scenario, is applicable when several datasets share the same sample ID space but differ in feature space. Consider, for instance, a bank and an e-commerce company in the same city. Their user sets are likely to contain most of the city's residents, so the intersection of their user base is large. However, their feature spaces are very different: the bank records information about a user’s revenue and expenditure behavior, while the e-commerce company retains the user’s browsing and purchasing history.
\end{itemize}

In this paper, we focus on VFL for classification problems, where the dataset also contains a finite number of labels and is heterogeneously distributed such that:
	\begin{itemize}
		\item Each node possesses its own portion of the input data but has no information about the other nodes’ inputs or the labels;
		\item The server exclusively owns the labels, while it is unaware of the input data.
	\end{itemize}
	
	In this setting, none of the federation’s components can perform independent training: the nodes cannot train without knowing the labels, and the server cannot train since it has no input data. Completing the VFL task, therefore, requires the collaborative effort of nodes and the server. This collaboration is organized in a sequence of \textit{training rounds}, each divided into several steps.
	\begin{itemize}
		\item[]\textbf{Step 1.} At the beginning of the round, each node uses its local model and input data to generate predictions on a batch, which are sent to the server.
		\item[]\textbf{Step 2.} The server aggregates these predictions and uses the result as input data. 
		\item[]\textbf{Step 3.} Since the server now has both inputs and labels, it can evaluate a given loss function and compute gradients to update its parameters. At the same time, the server also computes gradients of the loss function with respect to the node predictions it received, and sends these gradients to the nodes, enabling them to update their parameters. We emphasize that the server is the only federation member capable of computing these gradients, as it alone possesses the full information necessary to evaluate the loss function. 
		\item[]\textbf{Step 4.} The nodes use the gradients received to update their local parameters. The round is then concluded. 
	\end{itemize}
	
	Thus, VFL is a more complex process than standard centralized ML. Because of this complexity, three main aspects require special attention when designing a VFL algorithm:
	\begin{itemize}
		\item[] \textbf{Efficiency} – the algorithm should perform the classification task with limited computational resources and a minimal number of communications. 
		\item[] \textbf{Privacy preservation} – privacy must be preserved at all times to ensure that nodes have no access to data they do not own.
		\item[] \textbf{Generalization performance} – the trained federated model should generalize well, correctly classifying input data not used in its training.
	\end{itemize}
	
	These three aspects are often interlinked. High predictive performance may be achievable by allowing a large number of training rounds, but this comes at the cost of increased computational burden and vulnerability to adversarial attacks. Privacy may be enhanced by cryptographically encrypting shared data, but this raises the computational burden of the VFL process.
	
	As demonstrated in this work, our novel SBVFL scheme enhances efficiency while preserving privacy, maintaining high predictive performance compared to standard VFL. In Ren et al.~\cite{ren2022improving}, the concept of a soft label is introduced in a teacher-student VFL scheme. However, unlike synthetic labels, soft labels are essentially a continuous version of the real labels and may thus violate label privacy.
	
	The term \textit{blind} is also used in other technologies\footnote{For example, Fu et al.~\cite{fu2022blindfl} use HE or Secret Sharing (SS) to preserve privacy. However, these techniques further increase the already high communication overhead of VFL. In Salmeron et al.~\cite{salmeron2024blind}, the term \textit{blind} is used for an HFL scheme that does not require an initial model. An HFL use case is also presented by Razavikia et al.~\cite{razavikia2024blind}. These uses of \textit{blind} should not be confused with ours.}, but should not be confused with our usage.

\subsection{Mathematical Description of VFL}\label{subsection_standard_VFL}

We present a general mathematical setting for VFL here. This discussion aims to establish some notation and highlight the key aspects of VFL, which will later facilitate a comparison with our novel SBVFL methodology.

Given an input space $\mathcal X$, an output space $\mathcal Y$, and a dataset $\mathcal D = \left\{\left(\vec{x}^{\,i}, \vec{y}^{\,i}\right)\right\}_{i=1}^N\subset \mathcal{X}\times \mathcal{Y}$\footnote{This is the training dataset. Typically, in ML, in addition to this dataset, we have the validation $\mathcal{D}_{\mbox{\tiny{validation}}}=\left\{\left(\vec{x}_{\mbox{\tiny{validation}}}^{\,i},\vec{y}^{\,i}\right)\right\}_{i\in I_{\mbox{\tiny{validation}}}}$ and test $\mathcal{D}_{\mbox{\tiny{test}}}=\left\{\left(\vec{x}_{\mbox{\tiny{test}}}^{\,i},\vec{y}^{\,i}\right)\right\}_{i\in I_{\mbox{\tiny{test}}}}$ datasets.} composed of $N$ known but possibly noisy examples\footnote{\begin{definition}\label{def_noisy_example}
    Let $\left(\Omega,\mathscr{F},\mathbb{P}\right)$ be a probability space, $\left(\mathcal X, \mathscr{F}_x\right)$ be a measure space (typically, $\mathscr{F}_x$ is a Borel sigma-algebra) and $2^{\mathcal Y}$ be the power set of $\mathcal Y$.
    A pair $\left(\vec{x},\vec{y}\right)\in \mathcal{X}\times \mathcal{Y}$ is said to be a noisy example if it can be written as $\left(\vec{x},\vec{y}\right)=\left(\vec{x}_{\mbox{\tiny{truth}}},\vec{y}_{\mbox{\tiny{truth}}}\right)+\left(W_x,W_y\right)$, where $\left(\vec{x}_{\mbox{\tiny{truth}}},\vec{y}_{\mbox{\tiny{truth}}}\right)$ is the ground truth, $W_x:\left(\Omega,\mathscr{F}\right)\longrightarrow \left(\mathcal X, \mathscr{F}_x\right)$ is the random variable describing the noise on the $\mathcal X$ space and $W_y:\left(\Omega,\mathscr{F}\right)\longrightarrow \left(\mathcal Y, 2^{\mathcal Y}\right)$ is a random variable modeling noise in the labels set $\mathcal Y$ (see, e.g., Billings et al.~\cite{billingsley2012probability} for probability theory).
\end{definition}}, in a VFL scenario, a federation composed by $P$ nodes $\{\mathcal{F}_k\}_{k=1}^P$ and a server $\mathscr{S}$ aims to approximate an \textit{unknown function} $f:\mathcal X\to\mathcal Y$ mapping each input $\vec{x}^{\,i}\in \mathcal{X}$ to the corresponding output $\vec{y}^{\,i}\in \mathcal{Y}$ by collaboratively training a federated model 
\begin{align}\label{eq:fed_model}
	\mbox{FedMod}:\mathcal{X}\to\mathcal{Y}
\end{align}
such that 	
\begin{align*}
	\text{FedMod}(\vec{x}^{\,i}) \sim \vec{y}^{\,i}.
\end{align*}	

In this paper, the input space will always be $\mathcal{X}=\mathbb{R}^d$ with $d\geq 2$. Moreover, since we are focusing on classification problems, the output space is a finite set of classes 
\begin{align}\label{eq:labels}
	\mathcal{Y}\coloneqq\left\{\ell_1,\dots,\ell_M\right\}\subset \mathbb{R}^m,
\end{align}
with $m\geq 1$, $|\mathcal Y|=M$, and $\ell_p\neq \ell_q$ for all $p\neq q$\footnote{lowercase $p$ and $q$ index classes}. In \eqref{eq:labels}, we denote by $\{\ell_m\}_{m=1}^M$ the \textit{fixed} labels associated with the $M$ possible classes for our classification problem. This means that the labels $\{\vec y^{\,i}\}_{i=1}^N$ of our dataset all satisfy 
\begin{align*}
	\vec y^{\,i} = \ell_m, \text{ for some }\,m\in\inter{M}.
\end{align*}
Here, we have introduced the notation
\begin{align*}
	\inter{q}\coloneqq\{1,2,\ldots,q\} \text{ for all } q\in\mathbb{N},
\end{align*}
which we employ in the rest of the paper.

The main difference of VFL with respect to standard centralized ML is that the dataset $\mathcal D$ is decentralized among all the members of the federation. In more detail, each training example $\vec{x}^{\,i}\in \mathbb{R}^d$, $i\in\inter{N}$, is decomposed as 
\begin{align}\label{input_decomposition}
\vec{x}^{\,i}=\left(\vec{x}^{\,i}_1,\dots,\vec{x}^{\,i}_k,\dots,\vec{x}^{\,i}_P\right)
\end{align}
and the node $\mathcal{F}_k$ owns the $k$-th dataset 
\begin{align*}
	\mathcal D_k \coloneqq \{\vec{x}^{\,i}_k\}_{i=1}^N\subset\mathbb{R}^{d_k}, \quad\text{ with }\sum_{k=1}^{P}d_k = d. 
\end{align*}

The labels $\{\vec{y}^{\,i}\}_{i=1}^N$, instead, are exclusively owned by the server $\mathscr{S}$ and are kept unknown to the nodes. 

\begin{remark}\label{remark_active_party}
    In practical applications, the federation might be defined as follows.
    \begin{itemize}
        \item[i)] $P-1$ passive nodes $\mathcal{F}_1,\dots,\mathcal{F}_k,\dots,\mathcal{F}_{P-1}$, where
        each node $\mathcal{F}_k$ owns the unlabelled dataset
        \begin{align*}
        \mathcal D_k = \{\vec{x}^{\,i}_k\}_{i=1}^N\subset\mathbb{R}^{d_k} ;
        \end{align*}
        \item[ii)] an active node $\mathcal{F}_{\mbox{\tiny{A}}}$, possessing the labelled dataset
        \begin{align*}
        \mathcal D_{\mbox{\tiny{A}}} = \{\left(\vec{x}^{\,i}_P, \vec{y}^{\,i}\right)\}_{i=1}^N\subset\mathbb{R}^{d_P} \times \mathcal Y.
        \end{align*}
    \end{itemize}
    In this subsection notation, this means the \textit{physical active node} $\mathcal{F}_{\mbox{\tiny{A}}}$ contains \textit{two logical nodes}
    \begin{itemize}
        \item the passive node $\mathcal{F}_P$;
        \item the server $\mathscr{S}$.
    \end{itemize}
\end{remark}

Generally speaking, a strategy to obtain the federated model \eqref{eq:fed_model} relies on two main ingredients: 
\begin{itemize}
	\item[1.] The class of functions $\mathscr{C}$ to which FedMod should belong;
	\item[2.] The \textit{training method} to compute FedMod.
\end{itemize}

The choice of the class of functions $\mathscr{C}$ may depend on several factors, including prior knowledge of the dataset's intrinsic geometry and computational capabilities. Popular choices are Deep Neural Networks \cite{goodfellow2016deep}, Random Forest \cite{breiman2001random}, Gradient Boosting Decision Trees \cite{chen2016xgboost}, transformers \cite{vaswani2017attention}, Large Language Models \cite{naveed2023comprehensive} and Residual Neural Networks (ResNets) \cite{he2016deep}. The class considered will be denoted by
\begin{align*}
	\mathscr{C}=\big\{\text{FedMod}_{\*\theta} \ | \ \*\theta \in \*\Theta\big\},
\end{align*}
where $\*\theta$ is a vector containing all the model parameters belonging to some Hilbert space $\*\Theta$. In what follows, to simplify the notation, we omit the subscript $\*\theta$ when referring to the federated model FedMod, leaving its dependence on the model’s parameters understood. 

As for the training method, a common approach~\cite{chen2020vafl} is based on \textit{empirical risk minimization}. This consists of finding parameters $\*\theta^\ast$ solving
\begin{align}\label{opt_pb}
	\*\theta^\ast \in \underset{{\*\Theta}}{\mbox{argmin}} J(\*\theta),
\end{align}
with $J:\*\Theta\to \mathbb{R}$ defined, for instance, by 
\begin{equation}\label{eq:functional}
	J(\*\theta) \coloneqq \frac{1}{N}\sum_{i=1}^N \loss\Big(\text{FedMod}_{\*\theta}(\vec{x}^{\,i}), \vec{y}^{\,i}\Big) + \alpha \|\*\theta\|_{\*\Theta}^2,
\end{equation}
where the continuous loss function
\begin{align*}
	\loss:\mathbb{R}^d\times \mathcal{Y}\longrightarrow \mathbb{R}^+
\end{align*}
penalizes the mismatch between the predictions $\text{FedMod}_{\*\theta}(\vec{x}^{\,i})$ and the labels $\vec{y}^{\,i}$, while the \textit{regularization} term $\alpha \|\*\theta\|_{\*\Theta}^2$ penalizes the parameters complexity, $\|\cdot\|_{\*\Theta}$ being the Hilbertian norm. The effect of this penalization is modulated by the weighting factor $\alpha>0$.

We stress, however, that because of the heterogeneous data distribution in VFL, to solve the minimization problem \eqref{opt_pb}-\eqref{eq:functional}, neither the nodes nor the server can work independently, as they all miss some fundamental piece of information about the dataset. In fact, this minimization process requires the collaborative effort of nodes and the server. 

For this reason, $\text{FedMod}_{\*\theta}$ is made of the composition of nodes' models, each belonging to a function class chosen by each node.
In the following sections, we will provide a brief description of how this collaborative minimization is implemented. To this end, for all $k\in\inter{P}$, let us denote with 
\begin{align*}
	\text{Mod}_k:\mathbb{R}^{d_k}\to\mathbb{R}^{q_k},
\end{align*}
the local model of the node $\mathcal{F}_k$, $q_k\in\mathbb{N}\setminus\{0\}$ being the output dimension. We stress that, in general, this output dimension may be different from the local model's input dimension $d_k$. The model $\text{Mod}_k$ belongs to a class of functions
\begin{align*}
	\mathscr{C}_k=\big\{\text{Mod}_{k,\*\theta_k} \ | \ \*\theta_k \in \*\Theta_k\big\}.
\end{align*}
Analogously, also the server $\mathscr{S}$ possesses its own model
\begin{align*}
	\text{ServerMod}:\mathbb{R}^q\longrightarrow \mathbb{R}^{m},
\end{align*}
with $q=\sum_{k=1}^{P}q_k$, which is chosen in the class
\begin{align*}
	\mathscr{C}_{\scriptsize{s}}=\big\{\text{ServerMod}_{\*\theta_{\scriptsize{s}}} \ | \ \*\theta_{\scriptsize{s}} \in \*\Theta_{\scriptsize{s}}\big\}.
\end{align*}

Hence, the federated model on the full input $\vec{x}$ can be written in the compact form
\begin{equation}
\mbox{FedMod}(\vec{x})\coloneqq \text{ServerMod}\left(\text{Mod}_1(\vec{x}_1),\dots,\text{Mod}_P(\vec{x}_P)\right),
\end{equation}
the input $\vec{x}$ being decomposed as in \eqref{input_decomposition}.

As we did before, in what follows we will simplify the notation by dropping the subscripts $\*\theta_k$ and $\*\theta_{\scriptsize{s}}$, indicating the nodes' and server's model parameters.

As anticipated in Section \ref{sec:intro}~\cite{chen2020vafl}, the collaborative empirical risk minimization process is organized in a sequence of rounds, each divided into several steps. Let $I\subset \inter{N}$ be the indices associated with one batch.
\begin{itemize}
	\item[]\textbf{VFL training step 1.} At the beginning of the round, each node uses its local model and input data to generate predictions that are sent to the server. For every $i\in I$ and $k\in\inter{P}$, these predictions are obtained by computing 
	\begin{align*}
		\text{out}_k^{\,i}\coloneqq \text{Mod}_k(\vec{x}_k^{\,i})\in\mathbb{R}^{q_k}.	
	\end{align*}
	
	\item[]\textbf{VFL training step 2.} For all $i\in I$, the server aggregates the predictions $\{\text{out}_k^{\,i}\}_{k=1}^P$ into a concatenated vector 
	\begin{align*}
		\vec{x}_{\scriptsize{s}}^{\,i} \coloneqq \left[\text{out}_1^{\,i},\dots,\text{out}_P^{\,i}\right]\in\mathbb{R}^q,
	\end{align*}
	that is used as input for its model $\text{ServerMod}$. 
	\item[]\textbf{VFL training step 3.} Since the server now has both inputs and labels, it can evaluate 
	\begin{align*}
		\mbox{FedMod}(\vec{x}^{\,i}) = \text{ServerMod}(\vec{x}_{\scriptsize{s}}^{\,i})
	\end{align*}
	and the loss function in \eqref{eq:functional}. It can then compute gradients to update its parameters $\*\theta_{\scriptsize{s}}$ through
	\begin{equation}
		\*\theta_{\scriptsize{s}} \leftarrow \text{ServerUpdate}\Big(\*\theta_{\scriptsize{s}},\nabla_{\*\theta_{\scriptsize{s}}}J\Big).
	\end{equation} 
	Here, ServerUpdate is a gradient-based update procedure, such as one iteration of Stochastic Gradient Descent. At the same time, for all $i\in I$ and $k\in\inter{P}$, the server also computes $\nabla_{\text{out}_k^{\,i}}J$, i.e., the gradients of $J$ with respect to the node's predictions he received. These gradients are sent to the nodes, enabling their parameters to be updated. 
	\item[]\textbf{VFL training step 4.} In order to update its local parameters $\*\theta_k$, each node $\{\mathcal F_k\}_{k=1}^P$ should apply
	\begin{equation}
	\*\theta_k \leftarrow \text{NodeUpdate}\Big(\*\theta_k,\nabla_{\*\theta_k}J\Big),
	\end{equation}  
	where NodeUpdate is once again a gradient-based update procedure. However, none of the $\mathcal F_k$ is in a position to compute the above gradients, since they do not know the labels needed to evaluate $J$. Furthermore, not even the server $\mathscr S$ can compute the required gradients, as it has no information about the nodes' local parameters. To overcome this impasse, one may observe that, for all $i\in I$ and $k\in\inter{P}$,
	\begin{equation}
	\nabla_{\*\theta_k} J = \nabla_{\text{out}_k^{\,i}} J\partial_{\*\theta_k}[\text{out}_k^{\,i}]
	\end{equation}
	Now, each quantity $\partial_{\*\theta_k}[\text{out}_k^{\,i}]$ can be computed by the corresponding node $\mathcal F_k$, since it knows both its parameters and the output of its local model. As for $\nabla_{\text{out}_k^{\,i}} J$, this is exactly what the nodes receive from the server in Step 3. Thus, $\nabla_{\*\theta_k} J$ is obtained through the node-server collaboration. In this way, the local parameters $\{\*\theta_k\}_{k=1}^P$ can finally be updated. The round is then concluded. 
\end{itemize}

The above four-steps procedure clearly highlights one of the main issues in VFL: to train the federated model, it is required a very large number of communications between nodes and server, depending on the number of data, the batch size (if any) set in the optimizers ServerOptim and NodeOptim, the number of nodes $P$ and the total number of training epochs. In particular, we have
\begin{align}\label{eq:communications}
	\text{\# communications} \geq 2\left\lceil \frac{N}{\text{batch size}} \right\rceil P \Big(\text{\# epochs}\Big).
\end{align}

As observed by numerous studies (see, e.g., Chen et al.~\cite{chen2018lag} and the references therein), this high number of communications is often a bottleneck in overall VFL performance, both in terms of computational cost and privacy preservation. Our SBVFL paradigm, described in Section \ref{sec:BVFL}, has been designed specifically to address these issues, as it enables minimizing the number of nodes and server communications, resulting in a computationally efficient methodology with the added advantage of increased security compared to more traditional VFL approaches.

In Section \ref{sec:fake_labels}, we introduce a synthetic label generation mechanism that enables supervised learning on a dataset without exposing the ground-truth labels. Section \ref{sec:BVFL} then presents our SBVFL federated scheme. The synthetic label mechanism is the central novelty of SBVFL, yet it is broadly applicable to scenarios where labels are unavailable or cannot be directly shared.

\section{Synthetic Label Generation Mechanism}\label{sec:fake_labels}

The employment of synthetic labels during the federated training is a cornerstone of our SBVFL paradigm and one of the main differences with respect to other existing VFL methodologies. Synthetic labels can be thought of as intermediate goals, replacing final goals (real labels) for label-less nodes. Their generation is entrusted to the server, through a private procedure of which the nodes are unaware. Each node only receives the outcome of this procedure, and the server is the only one capable of associating these synthetic labels with the real ones. In this way, the privacy of real labels is preserved at all times.  

We describe in detail the process of generating these synthetic labels. In this procedure, $\mathscr S$ only needs to know the output dimensions $\{q_k\}_{k=1}^P$ of the nodes' models, and operates in three steps.
\begin{itemize}
	\item[] \textbf{Step 1.} Given the set of real labels 
	\begin{align*}
		\mathcal Y =\{\ell_1,\ldots,\ell_M\}\subset \mathbb{R}^m,
	\end{align*}
	the server, for each node $\mathcal F_k$, defines a (possibly nonlinear) matrix-valued operator
	\begin{align}\label{eq:Lambda_k_def}
		\Lambda_k:\mathcal{Y} \rightarrow \mathcal M^{Q \times q_k}(\mathbb{R}), \quad Q\in\mathbb{N}\setminus\{0\},
	\end{align}
	where $\mathcal M^{Q \times q_k}(\mathbb{R})$ denotes the space of real-valued matrices with dimension $Q\times q_k$. In what follows, to simplify the notation, we indicate this space simply with $\mathcal M^{Q \times q_k}$. This operator $\Lambda_k$ associates $Q$ synthetic labels of dimension $q_k$ to each $\ell_m$, $m\in\inter{M}$, these $Q$ synthetic labels being the rows $\{\mathbf{M}^{(j)}_m\}_{j=1}^Q$ of the matrix
	\begin{align*}
		\mathbf{M}_m\coloneqq\Lambda_k(\ell_m)\in \mathcal M^{Q \times q_k}. 
	\end{align*}
	
	\item[] \textbf{Step 2.} Associated with each $\Lambda_k$, the server also defines a \textit{privacy-preserving map} 
	Here, the number of rows $Q$ of the image of $\Lambda_k$ can be arbitrary. The more rows $Q$ we take, the more laborious the task of creating synthetic labels will be, but, simultaneously, the more secure the synthetic label generation will be. Indeed, let us assume for one moment that $Q=1$, i.e., to each real label we associate a unique synthetic one. For example, for a binary classification task, if the real label is $\{0,1\}$, the synthetic label could be $\{1.3, -0.7\}$ or $\{0.5, 2.1\}$ depending on $Q$. Then, the node $\mathcal F_k$ could cluster its examples, depending on the received synthetic label, and it would suffice that $\mathcal F_k$  knows the real label for just one example of each cluster to recover all the real labels. On the other hand, increasing $Q$, the number of distinct clusters for the same real label, also increases, making this recovery procedure harder. Due to its importance in preserving privacy during the generation of synthetic labels, we refer to $Q$ as the \textit{privacy multiplier}. 
	\begin{equation}\label{fake_labels_projection}
		\begin{array}{ll}
			\begin{array}{llll} \mathbb{P}_k: & \mathbb{R}^{q_k} & \to &\mathcal{Y} \\ & \mathbf{M}^{(j)}_m & \mapsto & \ell_m \end{array} 
			&
			\text{ for all } j\in\inter{Q}, 
		\end{array} 
	\end{equation}
	associating each synthetic label with its corresponding real one. Notice that this $\mathbb{P}_k$ can be seen as the inverse of the map $\Pi_j\circ\Lambda_k$, where for all $j\in\inter{Q}$, $\Pi_j$ denotes the projection to the $j$-th row of $\Lambda_k(\ell_m)$. Moreover, we stress that both $\mathbb{P}_k$ and $\Lambda_k$ are known only by the server $\mathscr S$.
	
	\item[] \textbf{Step 3.} Assume that $Q\geq 1$ has been fixed and that the operators $\{\Lambda_k\}_{k=1}^P$ have been built so that for all $m\in\inter{M}$ and $p,r\in\inter{Q}$
	\begin{align*}
		\mathbf{M}_m^{(p)}\neq \mathbf{M}_m^{(r)}, \quad\text{ if } p\neq r.
	\end{align*}
	Now, for all $i\in\inter{N}$ and $k\in\inter{P}$, the server generates the synthetic label $\vec z_k^{\,i}\in\mathbb{R}^{q_k}$, associated to the real label $\vec y^{\,i}$ and the node $\mathcal F_k$, by randomly selecting one of the rows of the matrix $\mathbf{M}_{m_i}=\Lambda_k(\ell_{m_i})$, where $\ell_{m_i}\in\mathcal Y$ denotes the label class of $\vec y^{\,i}$. 	
\end{itemize}

\begin{figure}[h]
	\centering
        \includegraphics[scale=0.4]{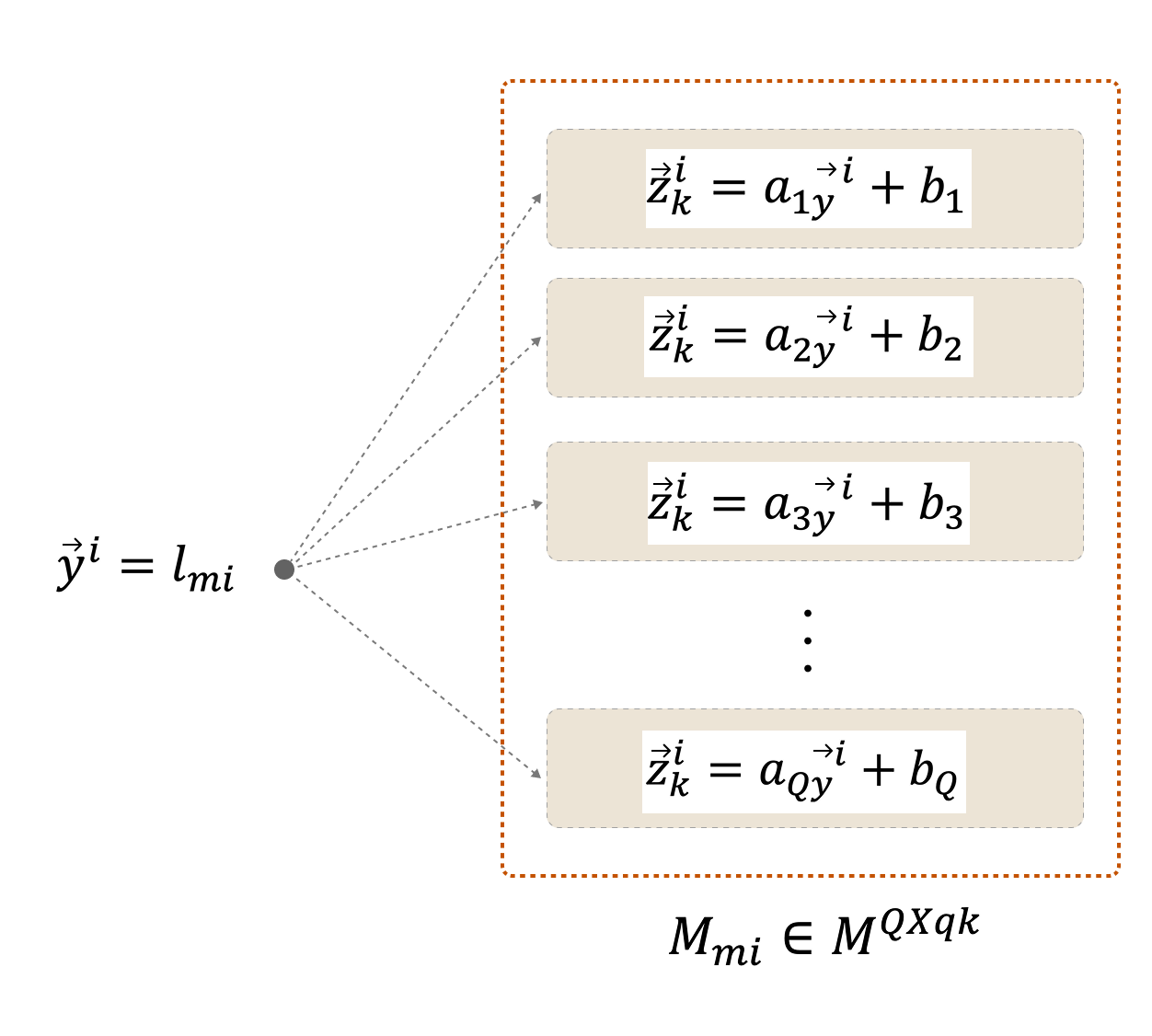}
	\caption{Synthetic labels generation through the flow of a hyperbolic PDE.}\label{fig:fake_labels}
\end{figure}

There are, of course, several ways to make the random selection at Step 3 of the procedure for generating synthetic labels that we have just described. Likewise, there are many ways to construct the maps $\{\Lambda_k\}_{k=1}^P$. In fact, any vector field could be employed. For example, we could define
\begin{align*}
	\vec z_k^{\,i} = \*a_j \vec y^{\,i} + \*b_j, \quad j\in\inter{Q},
\end{align*}
where for all $j\in\inter{Q}$, $\*a_j$ and $\*b_j$ are randomly chosen in $\mathcal M^{q_k\times n}$ and $\mathbb{R}^{q_k}$, respectively (see Figure \ref{fig:fake_labels}). This would be analogous to propagating the information of the real label $\vec y^{\,i}$ along the characteristics of a transport Partial Differential Equation (PDE). 

\section{The SBVFL Paradigm}\label{sec:BVFL}

\begin{figure}[h]
	\centering
        \includegraphics[width=1.0\columnwidth]{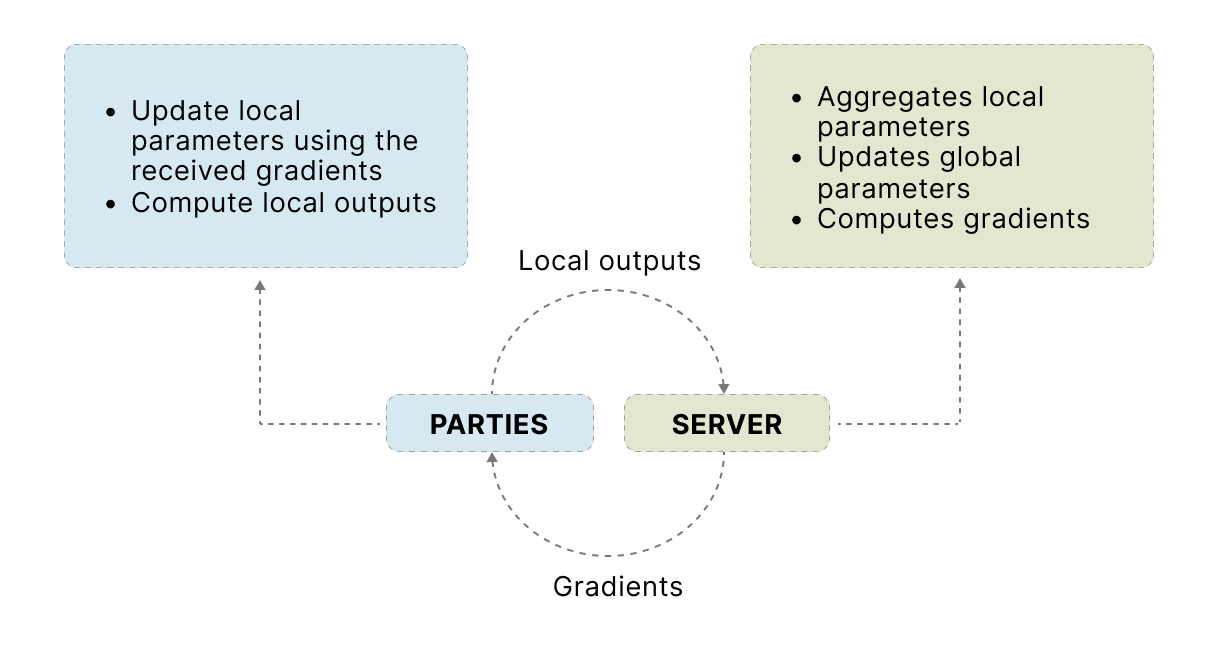}
	\caption{Scheme of the training rounds between nodes and server during VFL.}\label{fig:VFL_communications}
\end{figure}

\begin{figure}[h]
	\centering
        \includegraphics[scale=0.5]{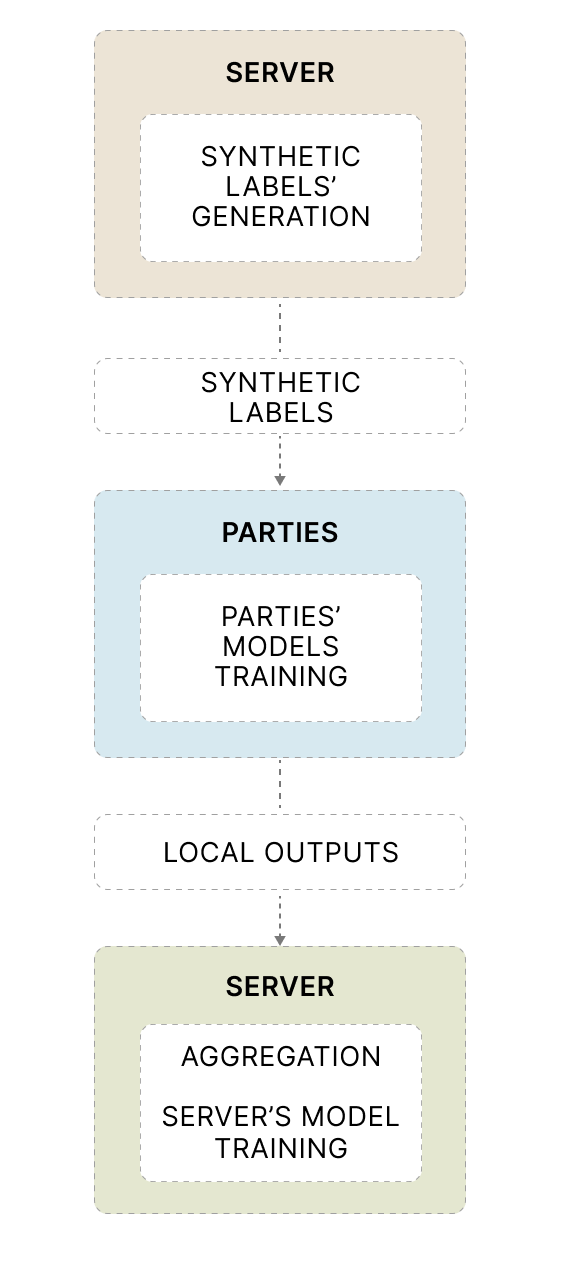}
	\caption{Scheme of the training process for SBVFL.}\label{fig:BVFL_scheme}
\end{figure}

As mentioned, the core idea behind SBVFL is to simplify VFL by employing synthetic labels, as described in Section \ref{sec:fake_labels}. This brings about a radical change in the federated training procedure, which, instead of being a \textit{circular} process organized into many communication rounds (see Figure \ref{fig:VFL_communications}), adopts the \textit{linear} structure illustrated in Figure \ref{fig:BVFL_scheme} without any further iteration.

In particular, in SBVFL, there are only two rounds of training, one for the nodes $\{\mathcal F_k\}_{k=1}^P$ in parallel and the other for the server $\mathscr S$. In this way, communication between nodes and the server is minimized, and the overall complexity of the training procedure is drastically reduced. In fact, under the SBVFL paradigm, training the federated model requires only one repetition of the following steps.

\begin{itemize}
	\item[] \textbf{SBVFL training step 0.} The server $\mathscr S$ generates synthetic labels that are distributed to all the nodes $\{\mathcal F_k\}_{k=1}^P$. 
	
	\item[] \textbf{SBVFL training step 1.} The nodes use their input data and the received synthetic labels to independently (and in parallel) train their local models $\{\text{Mod}_k\}_{k=1}^P$. 
	
	\item[] \textbf{SBVFL training step 2.} Once the local models are trained, the nodes in parallel generate predictions on their input data that are sent to the server. For every $i\in\inter{N}$ and $k\in\inter{P}$, these predictions are obtained by computing 
	\begin{align*}
		\text{out}_k^{\,i}= \text{Mod}_k(\vec{x}_k^{\,i}).
	\end{align*}
	
	\item[]\textbf{SBVFL training step 3.} For all $i\in\inter{N}$, $\mathscr S$ aggregates the predictions $\{\text{out}_k^{\,i}\}_{k=1}^P$ into a concatenated vector 
	\begin{align*}
		\vec{x}_{\scriptsize{s}}^{\,i} = \left[\text{out}_1^{\,i},\dots,\text{out}_P^{\,i}\right]\in\mathbb{R}^q,
	\end{align*}
        which is used as input for its model $\text{ServerMod}$. 
	
	\item[]\textbf{SBVFL training step 4.} $\mathscr S$ uses the aggregated input vectors $\{\vec{x}_{\scriptsize{s}}^{\,i}\}_{i=1}^N$ and the labels $\{\vec y^{\,i}\}_{i=1}^N$ to train its model.
\end{itemize}

This training procedure that we have just described is summarized in our SBVFL Algorithm \ref{alg:BVFL}.

\begin{remark}\label{rem:Light Vertical Federated Learning (LVFL)}
	The above architecture can be simplified by removing the SBVFL training steps $0$ and $1$. The resulting architecture is named in Sherpa.ai Light VFL (LVFL).
	
	Let us summarize the main steps below.
	\begin{itemize}
		\item[] \textbf{LVFL training step 0.} The passive nodes in parallel initialize their local models $\{\text{Mod}_k\}_{k=1}^P$ and generate predictions on their input data that are sent to the server. For every $i\in\inter{N}$ and $k\in\inter{P}$, these predictions are obtained by computing 
		\begin{align*}
		\text{out}_k^{\,i}= \text{Mod}_k(\vec{x}_k^{\,i}).
		\end{align*}
		
		\item[]\textbf{LVFL training step 1.} For all $i\in\inter{N}$, the server $\mathscr S$ aggregates the predictions $\{\text{out}_k^{\,i}\}_{k=1}^P$ into a concatenated vector 
		\begin{align*}
		\vec{x}_{\scriptsize{s}}^{\,i} = \left[\text{out}_1^{\,i},\dots,\text{out}_P^{\,i}\right]\in\mathbb{R}^q,
		\end{align*}
		that is used as input for its model $\text{ServerMod}$. 
		
		\item[]\textbf{LVFL training step 2.} $\mathscr S$ uses the aggregated input vectors $\{\vec{x}_{\scriptsize{s}}^{\,i}\}_{i=1}^N$ and the labels $\{\vec y^{\,i}\}_{i=1}^N$ to train its model.
	\end{itemize}
	The number of communications is equal to the number of nodes $P$.
	If, for each $k\in \left\{1,\dots,P\right\}$, the $\mathcal F_k$'s model $\text{Mod}_k:\mathbb{R}^{d_k}\to\mathbb{R}^{q_k}$ is one-to-one, it would suffice $\text{ServerMod}$ to be complex enough to have a successful classification.
	
	The drawback of LVFL is that the server training task may be computationally intensive, whereas in SBVFL, the passive nodes' training serves as a preprocessing step, which alleviates the server's task.
\end{remark}

\begin{algorithm}
	\vspace{0.2cm}
	\textbf{Input:} datasets $\mathcal{D}_k=\{\vec{x}^{\,i}_k\}_{i=1}^N$ owned by the node $\mathcal{F}_k$ for any $k\in\inter{P}$. Labels $\{\vec{y}^{\,i}\}_{i=1}^N$ owned by the server $\mathscr{S}$.
	
	\vspace{0.3cm}
	\textbf{Output:} trained federated model \eqref{eq:fed_model}, classifying the inputs $\{\vec{x}^{\,i}=\left(\vec{x}_1^{\,i},\dots,\vec{x}_P^{\,i}\right)\}_{i=1}^N$ to the corresponding labels $\{\vec{y}^{\,i}\}_{i=1}^N$.
	
	\vspace{0.3cm}
	\textbf{Procedure:}
	\begin{algorithmic}[1]
		\STATE For every $k\in \inter{P}$, the server $\mathscr{S}$ generates synthetic labels $\{\vec{z}_k^{\,i}\}_{i=1}^N\subset\mathbb{R}^{q_k}$ and sends them to node $\mathcal{F}_k$.
		
		\FOR{$k\in\inter{P}$ in parallel}
		
			\STATE Each node $\mathcal{F}_k$ trains its model $\text{Mod}_k$ via empirical risk minimization, by looking for the parameters $\*\theta_k^\ast$ solving
			\begin{align*}
				\*\theta_k^\ast = \underset{{\*\theta_k\in \*\Theta_k}}{\mbox{argmin}}\, J_k,
			\end{align*}
			with $J_k :\*\Theta_k\to \mathbb{R}$ given by 
			\begin{align*}
				J_k(\*\theta_k) = \frac 1N\sum_{i=1}^N \loss\Big(\text{Mod}_k(\vec{x}_k^{\,i}), \vec{z}_k^{\,i}\Big) + \alpha_k \|\*\theta_k\|_{\*\Theta_k}^2.
			\end{align*}
		
			\STATE $\mathcal{F}_k$ sends $\{\text{Mod}_k(\vec{x}_{k}^{\,i})\}_{i=1}^N$ to the server $\mathscr{S}$. \label{step:2}
		\ENDFOR	
			
		\STATE The server $\mathscr{S}$ aggregates the received output into the vectors $\{\vec x_{\scriptsize{s}}^{\,i}\}_{i=1}^N$ with
		\begin{align*}
			\vec x_{\scriptsize{s}}^{\,i}=[\text{Mod}_1(\vec{x}_{1}^{\,i}),\dots,\text{Mod}_P(\vec{x}_{P}^{\,i})]\in\mathbb{R}^q \text{ for all } i\in\inter{N}.
		\end{align*}
		
		\STATE The server $\mathscr{S}$ trains its model via empirical risk minimization, by looking for the parameters $\*\theta_{\scriptsize{s}}^\ast$ solving
		\begin{align*}
			\*\theta_{\scriptsize{s}}^\ast = \underset{{\*\theta_{\scriptsize{s}}\in \*\Theta_{\scriptsize{s}}}}{\mbox{argmin}} J_{\scriptsize{s}},
		\end{align*}
		with $J_{\scriptsize{s}} :\*\Theta_{\scriptsize{s}}\to \mathbb{R}$ given by
		\begin{align*}
			J_{\scriptsize{s}}(\*\theta_{\scriptsize{s}}) = \frac 1N\sum_{i=1}^N \loss\Big(\text{ServerMod}(\vec{x}_{\scriptsize{s}}^{\,i}), \vec{y}^{\,i}\Big) + \alpha_{\scriptsize{s}} \|\*\theta_{\scriptsize{s}}\|_{\*\Theta_{\scriptsize{s}}}^2.
		\end{align*}
	\end{algorithmic}
	\caption{SBVFL Training}\label{alg:BVFL}
\end{algorithm}

Finally, to conclude this section, let us highlight that to perform the federated training, our SBVFL Algorithm \ref{alg:BVFL} requires only $2P$ node-server communications, since the only information shared is the synthetic labels sent from the server to the nodes and the local training outputs sent from the nodes to the server. This represents a substantial reduction compared to traditional VFL (see \eqref{eq:communications}), which yields several benefits: it reduces computing time, costs, and carbon footprint~\cite{qiu2020planet}, and enhances privacy. Note that, in VFL, the privacy of the labels owned by the server can be in danger, as illustrated by Fu et al.~\cite{fu2022label}.	

\section{Theoretical Analysis of SBVFL}\label{sec:control}

To further substantiate SBVFL’s effectiveness, we provide a theoretical analysis that leverages control theory to demonstrate how SBVFL can achieve the same classification capability as a centralized model. 

To this end, we adopt the viewpoint of \cite{ruiz2023neural} and discuss the efficacy of our blind methodology by relating it to the simultaneous controllability properties of NeurODEs~\cite{agrachev2020control,cuchiero2020deep,tabuada2020universal}, a continuous version of ResNets. 

In particular, for all $k\in\inter{P}$, the local model $\text{Mod}_k$ of the node $\mathcal{F}_k$ is the following NeurODE
\begin{equation}\label{NODEs_3}
	\begin{cases}
		\dot{\*x}_k(t) \!= \*w_k(t)\sigma\left(\*a_k(t)\cdot\*x_k(t)\!+b_k(t)\right), & \!\!t\in (0,1)
		\\
		\*x_k(0)\!=\vec{x}_k\in \mathbb{R}^{d_k},
	\end{cases}
\end{equation}
with controls $\*a_k,\*w_k \in L^{\infty}(0,1;\mathbb{R}^{d_k})$ and $b_k\in L^{\infty}(0,1;\mathbb{R})$, to which it corresponds a flow
\begin{displaymath}
	\begin{array}{llll}
		\phi_k\left(\cdot;\*a_k,b_k,\*w_k\right): & \mathbb{R}^{d_k} & \to & \mathbb{R}^{d_k} 
		\\
		& \vec{x}_k & \mapsto & \*x_k(1).
	\end{array} 
\end{displaymath}

Here, the controls $\*a_k$, $b_k$, and $\*w_k$ play the role of the model's parameters $\*\theta_k$ in our previous discussion in Section \ref{sec:preliminaries}.

Analogously, also the server's model ServerMod is given by a NeurODE
\begin{equation}\label{NODEs_346}
	\begin{cases}
		\dot{\*x}_{\scriptsize{s}}(t) = \*w_{\scriptsize{s}}(t)\sigma\left(\*a_{\scriptsize{s}}(t)\cdot\*x_{\scriptsize{s}}(t)+b_{\scriptsize{s}}(t)\right), & \!\!t\in (0,1)
		\\
		\*x_{\scriptsize{s}}(0)=\vec{x}_{\scriptsize{s}},
	\end{cases}
\end{equation}
with associated flow
\begin{displaymath}
	\begin{array}{llll}
		\phi_{\scriptsize{s}}\left(\cdot;\*a_{\scriptsize{s}},b_{\scriptsize{s}},\*w_{\scriptsize{s}}\right): & \mathbb{R}^d & \to & \mathbb{R}^d 
		\\
		& \vec{x}_{\scriptsize{s}} & \mapsto & \*x_{\scriptsize{s}}(1).
	\end{array} 
\end{displaymath}

Within this framework, the federated model makes predictions on the input data by following a two-step procedure.
\begin{itemize}
	\item[] \textbf{Prediction step 1.} For all $i\in\inter{N}$ and $k\in\inter{P}$, the node $\mathcal{F}_k$ solves its NeurODE \eqref{NODEs_3} with initial datum $\vec{x}^{\,i}_k\in\mathbb{R}^{d_k}$ and sends the final-time solution $\*x_k^{\,i}(1)$ to the server $\mathscr{S}$. Notice that, this time, the output $\*x_k^{\,i}(1)$ being given by the flow of an ODE, the input and output dimensions of $\text{Mod}_k$ coincide ($q_k=d_k$). 
	
	\item[] \textbf{Prediction step 2.} The server $\mathscr{S}$ aggregates the received outputs into
	\begin{align*}
		\vec{x}_{\scriptsize{s}}^{\,i} = \left[\*x_1^{\,i}(1),\dots,\*x_P^{\,i}(1)\right]\in\mathbb{R}^d,
	\end{align*}
	solves its NeurODE \eqref{NODEs_346} with initial datum $\vec{x}_{\scriptsize{s}}^{\,i}$ and generates the prediction 
	\begin{align*}
		\mbox{FedMod}\left(\vec{x}^{\,i}\right)= \mathbb{P}[\*x_{\scriptsize{s}}^{\,i}(1)],
	\end{align*}
	where $\mathbb{P}$ is the projector
	\begin{equation}\label{Proj}
	\begin{array}{llll}
	\mathbb{P}: & \mathbb{R}^d & \to & \mathcal{Y}
	\\
	& \vec{z} & \mapsto & \sum_{r=1}^{M}\vec{y}^{\,[r]}\chi_{\Omega_r}\left(\vec{z} \hspace{0.08 cm} \right),
	\end{array} 
	\end{equation}
	defined by the partition $\{\Omega_r\}_{r=1}^M$ of $\mathbb{R}^d$
	\begin{equation}\label{Centr_Omega_partition}
	\mathbb{R}^d=\Omega_1\sqcup \dots \sqcup \Omega_M
	\end{equation}
	where $\Omega_r$ has nonempty interior for any $r\in \left\{1,\dots,M\right\}$. Note that the scope of the projector \eqref{Proj} is then twofold:
	\begin{itemize}
		\item dimensional projection from $\mathbb{R}^d$ to $\mathbb{R}^m$;
		\item mapping to the finite set $\mathcal{Y}$.
	\end{itemize}
\end{itemize}

The training of FedMod, instead, is carried out employing the simultaneous controllability algorithm defined in \cite[Theorem 1]{ruiz2023neural}, which essentially consists of a constructive procedure to compute controls $\{(\*a^*_k, b^*_k, \*w^*_k)\}_{k=1}^P$ and $(\*a^*_{\scriptsize{s}}, b^*_{\scriptsize{s}}, \*w^*_{\scriptsize{s}})$ such that 
\begin{align*}
	\mathbb{P}(\*x^{\,i}_{\scriptsize{s}}(1)) = \vec{y}^{\,i} \text{ for all } i\in\inter{N}. 
\end{align*}

We emphasize that, as observed in \cite{esteve2020large}, optimal control theory can be used to relate this training procedure to the empirical risk minimization process. In fact, a practical procedure to compute the controls $\{(\*a^*_k, b^*_k, \*w^*_k)\}_{k=1}^P$ and $(\*a^*_{\scriptsize{s}}, b^*_{\scriptsize{s}}, \*w^*_{\scriptsize{s}})$ consists precisely in minimizing the functionals 
$J_k(\*\theta_k)$ and $J_{\scriptsize{s}}(\*\theta_{\scriptsize{s}})$ introduced in Algorithm \ref{alg:BVFL}. Under standard regularity assumptions on the map
\begin{align*}
	\*\theta\mapsto \text{FedMod}_{\*\theta}(\vec{x}),
\end{align*}
for instance assuming it is of class $C^1$ and weakly continuous for any $\vec{x}\in \mathbb{R}^{d}$, this procedure can be proven to be successful 
by means of the Direct Method in the Calculus of Variations \cite{dacorogna2007direct}.

Finally, we can employ \cite[Theorem 1]{ruiz2023neural} to rigorously prove the efficiency of our SBVFL paradigm in solving the classification task for which it was designed. To this end, in what follows, we assume $d_k\geq 2$ for all $k\in\inter{P}$. This is required to apply the simultaneous controllability strategy of Ruiz et al.~\cite{ruiz2023neural}. We stress, however, that this assumption is not restrictive. Indeed, in case the data allocated to one node is in $\mathbb{R}$, we can easily embed them into $\mathbb{R}^2$ by applying, for instance, the map
\begin{displaymath}
\begin{array}{rccc}
	E: & \mathbb{R} & \to & \mathbb{R}^2
	\\
	& \xi & \mapsto &(\xi,0).
\end{array}
\end{displaymath}
With this said, we can prove the following result. 

\begin{theorem}\label{th_BVFL}
Suppose that $d_k\geq 2$ for all $k\in\inter{P}$. Assume that the centralized dataset $\left\{\left(\vec{x}^{\,i}, \vec{y}^{\,i}\right)\right\}_{i=1}^N$ has distinct input data $\vec{x}^{\,i}\neq \vec{x}^{\,j}$ for $i\neq j\in\inter{N}$. Then, the federated model $\emph{FedMod}$, trained by Algorithm \ref{alg:BVFL}, is able to classify each input into the corresponding label. Namely, 
\begin{align*}
	\emph{FedMod}\left(\vec{x}^{\,i}\right)=\vec{y}^{\,i}, \quad \emph{for all } i\in\inter{N}.
\end{align*}
\end{theorem}

\begin{proof}
	
The proof follows directly from the simultaneous controllability results for NeurODEs obtained by Ruiz et al.~\cite{ruiz2023neural}. In fact, we only have to show that the server $\mathscr S$, after having received from the nodes the outputs $\{\phi_k(\vec{x}_k^{\,i})\}$ for all $i\in\inter{N}$ and $k\in\inter{P}$, is capable of classifying the aggregated vectors
\begin{align*}
	\vec x_{\scriptsize{s}}^{\,i}=[\phi_1(\vec{x}_{1}^{\,i}),\dots,\phi_P(\vec{x}_{P}^{\,i})]\in\mathbb{R}^d
\end{align*}
by computing controls $\*a_{\scriptsize{s}}$, $b_{\scriptsize{s}}$ and $\*w_{\scriptsize{s}}$ such that
\begin{align*}
	\phi_{\scriptsize{s}}(\vec x_{\scriptsize{s}}^{\,i}) = \vec{y}^{\,i}, \quad \text{for all } i\in\inter{N}.
\end{align*} 
	
To this end, it is enough to show that the input data of the server are all different, that is, 
\begin{align}\label{eq:server_diff}
	\vec{x}_{\scriptsize{s}}^{\,i} \neq \vec{x}_{\scriptsize{s}}^{\,j} \quad\text{ for all } i\neq j\in\inter{N}.
\end{align}
	
Indeed, once \eqref{eq:server_diff} is satisfied, we can apply the constructive procedure of \cite[Theorem 1]{ruiz2023neural} to determine the desired controls. 
	
Now, since by assumption $\vec{x}^{\,i}\neq \vec{x}^{\,j}$ for all $i\neq j\in\inter{N}$, there exists at least one $k\in\inter{P}$ such that
\begin{align*}
	\vec{x}^{\,i}_k \neq \vec{x}^{\,j}_k.
\end{align*}
Therefore, using the uniqueness of solutions to \eqref{NODEs_3}, we get
\begin{align*}
	\phi_k(\vec{x}_{k}^{\,i})\neq \phi_k(\vec{x}_{k}^{\,j}),
\end{align*}
from which \eqref{eq:server_diff} follows immediately. Our proof is then concluded.
\end{proof}

\begin{remark}
\em{
In light of remark \ref{rem:Light Vertical Federated Learning (LVFL)}, the conclusions of Theorem \ref{th_BVFL} hold if SBVFL is replaced with LVFL.

Indeed, the proof only relies on the one-to-one property of the passive nodes’ models and the controllability properties of the server model.
}
\end{remark}

\section{Privacy and Security Analysis of SBVFL}\label{sec:privacy}

When presenting SBVFL in Section \ref{sec:BVFL}, we have claimed that our approach enhances privacy in VFL. The aim of this section is to provide a more concrete ground for these claims.

To this end, we consider a scenario in which one of the members of the federation, either a node or the server, is a malicious agent that wants to 
acquire information about data that is not in its possession. This can be translated into three possible types of attacks.
\begin{itemize}
	\item[] \textbf{Node-to-server attack:} one of the nodes $\mathcal F_k$ wants to recover the data of the server $\mathscr S$.
	\item[] \textbf{Server-to-node attack:} the server $\mathscr S$ wants to recover the data of one of the nodes $\mathcal F_k$.
	\item[] \textbf{Node-to-node attack:} one of the nodes $\mathcal F_k$ wants to recover the data of another node $\mathcal F_j$.
\end{itemize}

In what follows, we adopt the NeurODE perspective from Section \ref{sec:control} to investigate whether these attacks can be successful under the SBVFL protocol that we have designed.

\subsection{Node-to-server Attacks}

In a node-to-server attack, one of the nodes $\mathcal F_k$ tries to recover the data in possession of the server $\mathscr S$, that is, the labels $\{\vec y^{\,i}\}_{i=1}^N$ (see Fu et al.~\cite{fu2022label}
 for label inference attacks against VFL). 

Let us recall that, in the VFL scenario we are considering in this paper, the attacker $\mathcal F_k$ knows nothing about the labels, except for the synthetic labels it possesses. Hence, the only information this node can use to launch its attack is the synthetic labels it received from the server at the beginning of training. Nevertheless, this information is not enough to successfully complete the attack. Indeed, from the viewpoint of $\mathcal F_k$, these synthetic labels are simply some vectors $\vec z_k^{\,i}$, whose correlation with the real labels is made possible only through the projection $\mathbb{P}_k$ defined in \eqref{fake_labels_projection}, that is known only by the server.

Notwithstanding, we stress that, in some real-life scenarios, the attacker $\mathcal F_k$ may know some labels $\{\vec y^{\,i}\}_{i\in I}$, where $I\subset \inter{N}$. In this case, whenever the attacker receives a synthetic label $\vec z_k^{\,j}$, with
\begin{align*}
	\vec z_k^{\,j}=\vec z_k^{\,i},\hspace{0.3 cm}\mbox{for some }i\in I,
\end{align*}
it can deduce that the label for the local input $\vec{x}^{\,j}_k$ is $\vec y^{\,i}$. These attacks can be fully neutralized by generating distinct synthetic labels.
\begin{align*}
	\vec{z}_k^{\,i}\neq \vec{z}_k^{\,j}, \quad\text{ for all } i\neq j\in\inter{N}.
\end{align*}

In summary, our SBVFL paradigm is secure against node-to-server attacks, provided that synthetic label generation is performed appropriately. On the other hand, the classical VFL approach may be vulnerable to these kinds of attacks, as demonstrated by Fu et al.\cite{fu2022label}.

\subsection{Server-to-node Attacks}\label{subsec:server-party}

In a server-to-node attack, the server $\mathscr S$ wants to recover the data of one of the nodes $\mathcal F_k$, that is, the local dataset $\{\vec x_k^{\,i}\}_{i=1}^N$. To do that, $\mathscr S$ needs to use the final value $\*x_k(1)$ of the victim's NeurODE
\begin{equation}\label{NODEs_10}
\dot{\*x}_k(t) = \*w_k(t)\sigma\left(\*a_k(t)\cdot\*x_k(t)+b_k(t)\right), \;\;\;t\in (0,1).
\end{equation}

The adversary $\mathscr S$ is not aware of the specific controls $\*a_k$, $b_k$, and $\*w_k$, but knows the architecture of \eqref{NODEs_10}, namely the space dimension $d_k$ and the activation function $\sigma$. Besides, let us assume that $\mathscr S$ also knows that each example in the secret $\{\vec{x}^{\,i}_k\}_{i=1}^N$ must obey some constraints. We model this assuming the existence of a subset $\mathscr{V}\subseteq \mathbb{R}^{d_k}$, such that $\vec{x}^{\,i}_k\in \mathscr{V}$, for any $i \in \left\{1,\dots,N\right\}$. We can then formulate the attack of $\mathscr S$ to $\mathcal{F}_k$ in the following way.

\begin{definition}[$\mathcal{F}_k\leftarrow \mathscr S$ attack]\label{def:attack}
Let $\mathscr{V}\neq\{\*0\}$ be a non-empty subset of $\mathbb{R}^{d_k}$. Given the final values $\left\{\*x_k^{\,i}(1)\right\}_{i=1}^{N}$ of the victim's Neural ODE \eqref{NODEs_10}, find all possible datasets $\{\vec{x}_{k,\mbox{\tiny{sh}}}^{\,i}\}_{i=1}^{N} \subset \mathscr{V}$ such that there exist controls $\*a_{\mbox{\tiny{sh}}},\*w_{\mbox{\tiny{sh}}}\in L^{\infty}(0,1;\mathbb{R}^{d_k})$ and $b_{\mbox{\tiny{sh}}}\in L^{\infty}(0,1;\mathbb{R})$ for which the solution to the \emph{shadow} NeurODE
	\begin{equation}\label{NODEs_11}
	\begin{dcases}
	\dot{\*x}_{\mbox{\tiny{sh}}}^{\,i}(t) \!= \*w_{\mbox{\tiny{sh}}}(t)\sigma\left(\*a_{\mbox{\tiny{sh}}}(t)\cdot\*x_{\mbox{\tiny{sh}}}^{\,i}(t)+b_{\mbox{\tiny{sh}}}(t)\right), & \!\!\!t\in (0,1)
	\\
	\*x_{\mbox{\tiny{sh}}}^{\,i}(0) \!=\vec{x}_{k,\mbox{\tiny{sh}}}^{\,i}
	\end{dcases}
	\end{equation}
	satisfies the final condition
	\begin{equation}\label{eq:attack}
	\*x_{\mbox{\tiny{sh}}}^{\,i}(1)=\*x_k^{\,i}(1),
	\end{equation}
	for all $i\in\inter{N}$.
\end{definition}

According to this Definition \ref{def:attack}, an adversarial server-to-node attack essentially consists in solving an inverse problem: knowing the final states of \eqref{NODEs_10}, $\mathscr S$ uses \eqref{NODEs_11} to try discovering the $\mathcal{F}_k$'s secret $\{\vec{x}^{\,i}_k\}_{i=1}^N$ by computing all the possible initial data leading to those final configurations. Under this optics, an attack would then be unsuccessful whenever \eqref{NODEs_11} admits multiple solutions satisfying \eqref{eq:attack}, in which case the attacker $\mathscr S$ cannot identify the secrets $\vec{x}^{\,i}_k$. This motivates the following definition of \textit{$\mathcal{F}_k$ security}.

\begin{definition}[$\mathcal{F}_k$ security]\label{def_privacy_F_2_F_1}
We say that node $\mathcal{F}_k$ is secure in regard to the server $\mathscr S$ if the adversarial attack of Definition \ref{def:attack} admits multiple solutions, i.e., if there exist at least two distinct shadow datasets leading to the given outputs.
\end{definition}

We have the following result of privacy preservation, in the spirit of Definition \ref{def_privacy_F_2_F_1}.
\begin{theorem}\label{thm_privacy}
Let $\mathscr{V}\neq\{\*0\}$ be a non-empty subset of $\mathbb{R}^{d_k}$, with cardinality $|\mathscr{V}|\geq 2$. Then, for all $k\in\inter{P}$, node $\mathcal{F}_k$ is secure in regard to $\mathscr S$.
\end{theorem}

\begin{proof}
Take any shadow dataset $\mathscr{D}_{\mbox{\tiny{sh}}}\coloneqq \{\vec{x}_{k,\mbox{\tiny{sh}}}^{\,i}\}_{i=1}^{N}\subset \mathbb{R}^{d_k}$, with
\begin{align}\label{cond_1}
	\vec{x}_{k,\mbox{\tiny{sh}}}^{\,i}\neq \vec{x}_{k,\mbox{\tiny{sh}}}^{\,j} \hspace{0.3 cm}\mbox{whenever} \ \*x_k^{\,i}(1)\neq\*x_k^{\,j}(1).
\end{align}
	
Then, by means of the work done by Ruiz et Zuazua~\cite[Theorem 2]{ruiz2023neural}, there exist controls $\*a_{\mbox{\tiny{sh}}},\*w_{\mbox{\tiny{sh}}}\in L^{\infty}(0,1;\mathbb{R}^{d_k})$ and $b_{\mbox{\tiny{sh}}}\in L^{\infty}(0,1;\mathbb{R})$, such that \eqref{NODEs_11} admits a solution.
	
Hence, by Definition \ref{def:attack}, the shadow dataset $\mathscr{D}_{\mbox{\tiny{sh}}}$ is a solution to the adversarial attacks. Nevertheless, the choices of $\mathscr{D}_{\mbox{\tiny{sh}}}$ fulfilling \eqref{cond_1} are multiple, since at least the first example $\vec{x}_{k,\mbox{\tiny{sh}}}^{\,1}$ can be chosen arbitrarily in $\mathscr{V}$ (which, by assumption, has at least two elements). This means that there are (at least) two solutions to the adversarial attack and, therefore, the node $\mathcal F_k$ is secure in the sense of Definition \ref{def_privacy_F_2_F_1}.
\end{proof}

Theorem \ref{thm_privacy} can now be used to assess the security of our proposed SBVFL approach against server-to-node attacks. In fact, thanks to the employment of synthetic labels, for all $i\in\inter{N}$ the only information the server $\mathscr S$ has about the node $\mathcal F_k$ is the final state $\*x^{\,i}_k(1)$. According to Theorem \ref{thm_privacy}, this is not enough for $\mathscr S$ to break the security of $\mathcal F_k$.

In this regard, in practical applications, NeurODEs are most often replaced by some discretized counterpart like the so-called Residual Neural Network
\begin{align*}
\begin{dcases}
\*x^{t+1} = \*x^t + \*w^t\sigma\left(\*a^t\cdot\*x^t+b^t\right), &t \in \{0, \ldots, \mathcal T\}
\\
\*x^0=\vec{x}_0,
\end{dcases}
\end{align*}

In this setting, the control strategy defined by Ruiz et al.~\cite[Section 4]{ruiz2023neural} can still be applied, provided that the controls satisfy some constraints associated with the number $\mathcal T$ of layers in the neural network. In view of this, simultaneous controllability holds only for special initial and final data, which are distributed according to a sufficiently simple geometry. The greater the number of layers $\mathcal T$, the larger is the space of initial data that are simultaneously controllable. This, clearly, also has effects on privacy, which is enhanced by the model's complexity.

\subsection{Node-to-node Attacks}

In a node-to-node attack one of the nodes $\mathcal F_k$ wants to recover the data of another node $\mathcal F_j$, that is, the local dataset $\{\vec x_j^{\,i}\}_{i=1}^N$. Nevertheless, the only communications among $\mathcal F_k$ and $\mathcal F_j$ happen via the server $\mathscr S$, and we have already shown in Subsection \ref{subsec:server-party} that through server-to-node communications, adversarial attacks are impossible. Hence, SBVFL is a secure environment even against node-to-node attacks.

\section{Experiments on Benchmark Dataset}\label{sec:experimental_results}

We present the results of some simulation experiments performed to test our proposed SBVFL methodology. In particular, we compare the performance of SBVFL with a centralized ML scenario, in which all data are gathered in a single location, as well as with the standard VFL methodology presented in Section \ref{sec:preliminaries}. 

\subsection{Dataset and Models}
We applied SBVFL to a wide variety of use cases. In this section, we present an application for image classification. In Section \ref{sec:Real world use cases and benefits}, we illustrate an application to banking-insurance.

To this end, we simulate a federation of seven nodes $\{\mathcal F_k\}_{k=1}^7$ and a server $\mathscr S$ collaborating to solve an image classification task. The dataset for our experiments will be EMNIST Digits~\cite{cohen2017emnist}, composed of a total of $280,000$ handwritten digit characters ranging from $0$ to $9$
, and the corresponding $10$ balanced classes of labels.

\begin{figure}[htp]
	\begin{center}
		\includegraphics[width=6cm]{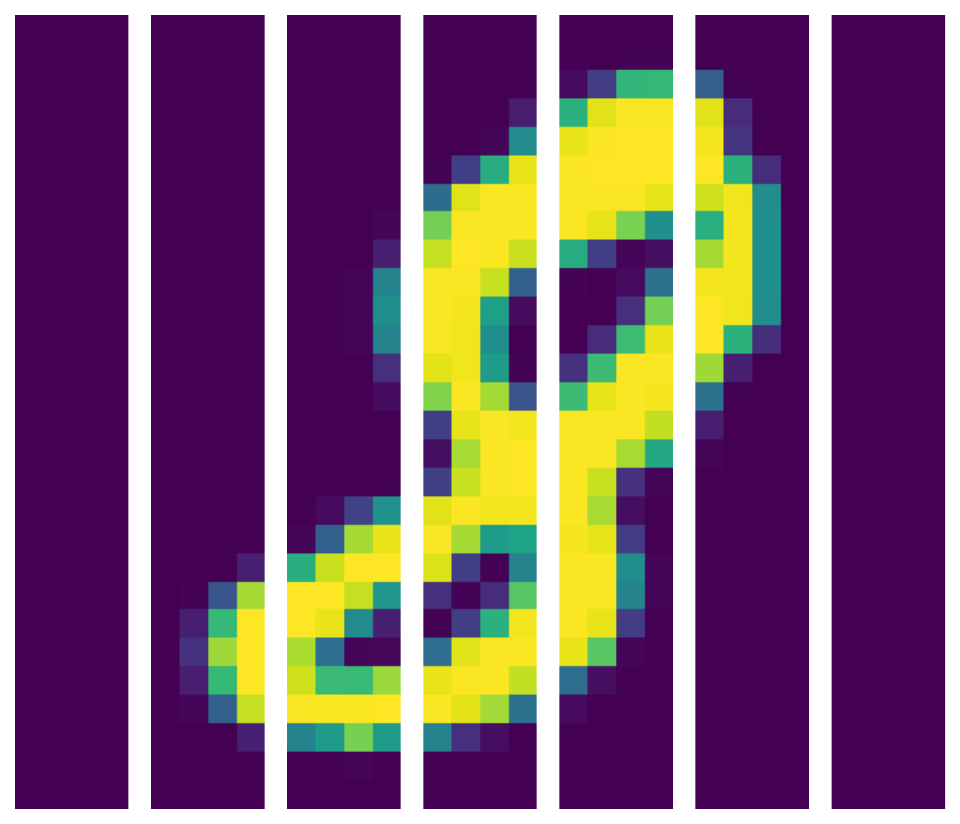}
		\caption{Example of EMNIST Digits image split in vertical slices, to be distributed among the nodes $\{\mathcal F_k\}_{k=1}^7$.}\label{fed_images}
	\end{center}
\end{figure}

Out of these $280,000$ images, $240,000$ will be used for training and the remaining $40,000$ ones will be reserved for testing. Moreover, to adapt the dataset to the federated scenario we are considering, we have performed a partition of the EMNIST Digits images into seven vertical slices (see Figure \ref{fed_images}), and assigned each one of these slices to one of the nodes $\{\mathcal F_k\}_{k=1}^7$. The labels, instead, are given to the server $\mathscr{S}$. Similar experiments (but with a different number of nodes) have been performed in \cite{romanini2021pyvertical}. In this manuscript, the number of nodes $P=7$ is chosen to ensure a sufficiently large gap between one-node-only performances and centralized performances.

Finally, as ML models for our training, we have used ResNets (see Ruiz et al.~\cite[Formula (1.2)]{ruiz2023neural}). Unless stated otherwise, all experiments were conducted using the following hyperparameters: a batch size of 128, a learning rate of $10^{-3}$, a weight decay coefficient of $10^{-3}$, a hidden dimension of 32, and a network depth of 10 layers.

\subsection{Experimentation Testbed}
Our experiments were conducted on a Dell computer equipped with a 3.6 GHz Intel Core i7-7700 processor and 32GB of RAM, using Python 3.8.

\subsection{Results}
\begin{figure}[h]
	\centering
        \includegraphics[width=1.0\columnwidth]{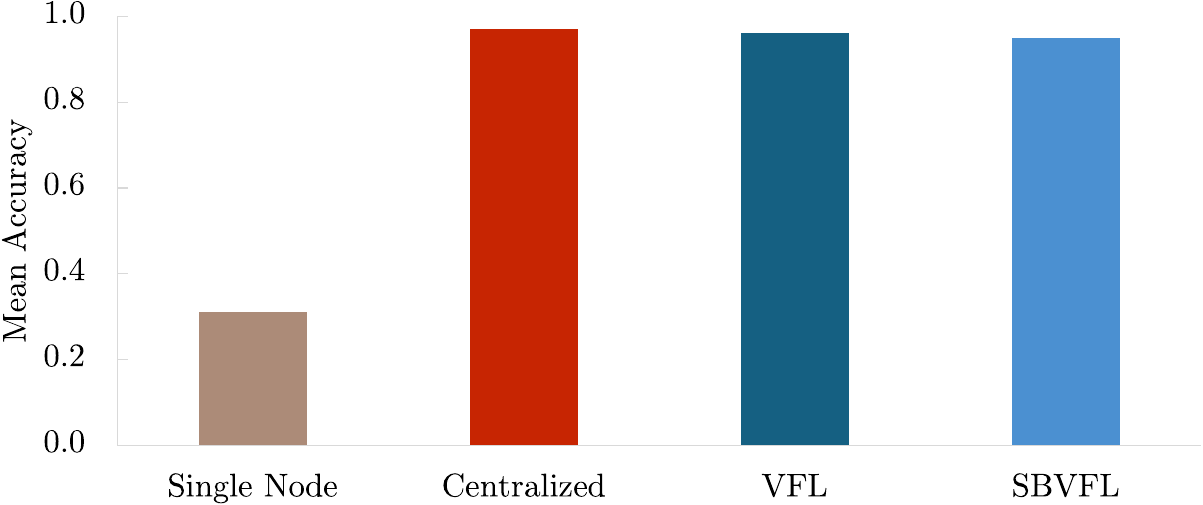}
	\caption{Mean accuracy for twenty random trials of Single Node, centralized ML, VFL, and SBVFL. }\label{fig:accuracy_100}
\end{figure}

In Figure~\ref{fig:accuracy_100} and Table \ref{tab:table_accuracy}, we display the mean accuracy for twenty experiments of centralized ML, VFL, and SBVFL. Notice that the centralized case is displayed for benchmarking purposes. However, it is not possible in practice, since it violates privacy regulations. In Table \ref{tab:table_accuracy}, the standard deviation (STD) of these experiments is also displayed.

We evaluated four training and evaluation configurations.
\begin{enumerate}
    \item \textbf{Single node}: Each node $\mathcal{F}_k$ trains its own model locally on its partition of the data; inference is performed on the held-out test set using the corresponding local model.
    \item \textbf{Centralized ML}: Data from all nodes are pooled to train a single model; inference on the test set uses this centralized model.
    \item \textbf{Standard VFL}: A federated model is trained jointly across all nodes; inference on the test set uses the resulting federated model.
    \item \textbf{SBVFL}: A federated model is trained jointly across all nodes under the SBVFL protocol; inference on the test set uses the resulting federated model.
\end{enumerate}



\begin{table}[h]
  \centering
  \begin{tabularx}{\columnwidth}{Y|Y|Y|Y|Y}
    \toprule
    \textbf{Metric} & \textbf{Single node (STD)} & \textbf{Centralized (STD)} & \textbf{VFL (STD)} & \textbf{SBVFL (STD)} \\
    \midrule
    \textbf{Accuracy} & 0.3105 (0.006) & 0.97 (0.0015) & 0.96 (0.0053) & 0.95 (0.0053)  \\
    \bottomrule
  \end{tabularx}
  \vspace{10pt}
  \caption{Mean accuracy and Standard Deviation (STD) for twenty random trials of centralized ML, VFL, and SBVFL.}
\label{tab:table_accuracy}
\end{table}

We can clearly observe that both VFL and SBVFL are capable of obtaining high accuracy, at a level that is close to the centralized scenario. Moreover, the low STD values (Table \ref{tab:table_accuracy}) we have obtained certify the stability of our experimental results over multiple repetitions.

\begin{figure}[h]
	\centering
        \includegraphics[width=1.0\columnwidth]{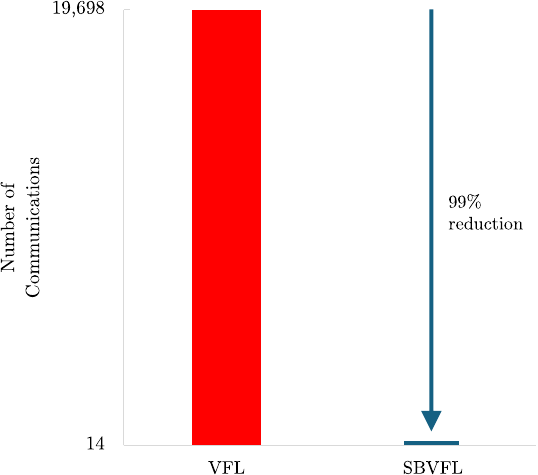}
	\caption{Total number of communications between the nodes and the server to complete the training in VFL and SBVFL. 
    }\label{fig:rounds}
\end{figure}

Notwithstanding, the real gain of SBVFL with respect to standard VFL is appreciated in the limited number of node-server communications that our approach needs to complete a federated training. This is clearly observed in Figure \ref{fig:rounds}, where we can see that our blind methodology allows for a $\sim$99\% reduction of these communications, whose number drastically drops from 19,698 to only 14 (that is, twice the number of nodes used in our experiments).


This is undoubtedly the most significant advancement of SBVFL, which, as discussed in the previous sections, leads to several benefits, including reduced computing time, costs, and carbon footprint, as well as enhanced privacy. 

After validating our approach on image data, we next demonstrate SBVFL on a real-world financial dataset in Section \ref{sec:Real world use cases and benefits}.

\section{Real-World Case Study (Financial Default Prediction)}\label{sec:Real world use cases and benefits}

Sherpa.ai’s FL technology is already being applied across a wide range of industries, including healthcare, banking and insurance, Industry 4.0, aerospace, and cybersecurity/defense—thanks to its strong focus on data privacy and security, readiness for enterprise deployments, and user-friendliness.

SBVFL achieves a communication reduction of $\sim$99\%, delivering significant advantages in scenarios where:

\begin{enumerate}
	\item Data security and privacy are paramount;
	\item Data transfer between distributed sources is costly or infeasible, such as with low-bandwidth satellite communications;
	\item Energy efficiency is critical due to hardware limitations, like constrained battery capacity.
\end{enumerate}

These conditions are especially relevant to make predictions in sectors such as banking and insurance, industrial environments, aerospace, and highly sensitive domains like cybersecurity and national defense. 


In this use case, we evaluate model performance across various training scenarios involving a bank and an insurance company, focusing on their ability to predict customer behavior while adhering to data privacy regulations. The scenarios range from no data sharing (single node training), where the bank trains its model independently, to full data sharing (centralized training), where both institutions pool their data. FL provides an optimal balance, achieving high predictive accuracy while preserving data privacy. For the FL approach, we implement two methods: standard VFL and SBVFL.

We have used four configurations for the process of training and evaluation:

\begin{enumerate}
    \item \textbf{Single Node}: The bank trains a separate model locally using its partitioned data. Inference is conducted on the test subset using the local model.
    \item \textbf{Centralized}: Data from both nodes is pooled into a single repository to train a unified model. Inference is performed on the test subset using the Centralized model. 
    \item \textbf{Standard VFL}: the model is trained with data from both nodes. Inference is performed on the test subset using the federated model. The data split across the nodes is detailed in the following section.
    \item \textbf{SBVFL}: the model is trained with data from both nodes. Inference is performed on the test subset using the federated model. The data split across the nodes is detailed in the following section.
\end{enumerate}

These models have been trained using the Sherpa.ai platform and evaluated over a centralized held-out test dataset. 

\subsection{Problem Formulation}

In the domain of consumer finance, predicting the likelihood of credit card default is a crucial task for financial institutions seeking to mitigate risk and maintain portfolio stability. The problem addressed herein is classifying credit card customers into two categories—defaulters and non-defaulters—based on historical and demographic data.

The objective is to develop a predictive model that accurately estimates the probability of a customer defaulting on their credit card payment in the upcoming month. The dataset used for this task consists of 30,000 observations of credit card customers in Taiwan, collected by a major financial institution. Each observation includes 23 explanatory variables, encompassing customer-specific features such as
\begin{itemize}
    \item demographic attributes (e.g., age, gender, education, and marital status),
    \item financial data (e.g., credit limit, payment history, and bill statement amounts), and
    \item behavioral data over the preceding six months (e.g., payment amount and delay status).
\end{itemize}

The target variable \verb!default payment next month! is a binary indicator of default payment in the next month (1 = default, 0 = non-default).

In the context of Section \ref{subsection_standard_VFL}, in the centralized case, we would minimize a functional
\begin{equation}\label{eq:functional_defaulters}
	J(\*\theta) \coloneqq \frac{1}{N}\sum_{i=1}^N \loss\Big(\text{Mod}_{\*\theta}(\vec{x}^{\,i}), \vec{y}^{\,i}\Big) + \alpha \|\*\theta\|_{\*\Theta}^2,
\end{equation}
where
\begin{itemize}
    \item the input space is $\mathcal{X}=\mathbb{R}^{23}$ (the dimension $d=23$);
    \item the labels set is $\mathcal{Y}=\left\{0, 1\right\}$ (binary classification).
\end{itemize}

\underline{Goal}: learn a mapping 
\begin{equation}
f: \mathbb{R}^{23} \longrightarrow \{0, 1\}
\end{equation}
such that \( f(\vec{x}^{\,i}) \) predicts whether the customer \( i \) will default.

This classification problem can be addressed using various ML models, including Deep Neural Networks~\cite{goodfellow2016deep}, Random Forests~\cite{breiman2001random}, Gradient Boosting Decision Trees~\cite{chen2016xgboost}, Large Language Models~\cite{naveed2023comprehensive}, and Residual Neural Networks (ResNets)~\cite{he2016deep}. The dataset is imbalanced (the number of defaulting customers is $22.12\%$ of the total). Therefore, metrics for imbalanced datasets, such as the area ratio (as in the seminal paper by Yeh and Lien~\cite{YEH20092473}), F1-score, or AUC, should be adopted to assess model prediction performance.

\subsection{Description of the dataset}

For this experiment, we use the Default of Credit Card dataset~\cite{YEH20092473}, accessible via the UCI ML Repository. This dataset, collected from 30,000 Taiwanese credit card holders between April and September 2005, includes 23 features such as demographic details (age, gender, education, marital status), credit information (credit limit, payment history), and six months of billing and payment records.

The original study aimed to assess the predictive performance of various data mining methods for forecasting default payment probabilities. This dataset is a robust resource for classification tasks in credit risk modeling and financial analytics.


\subsection{Federated Setup}

\begin{figure}[h]
    \centering
    \includegraphics[width=\columnwidth]{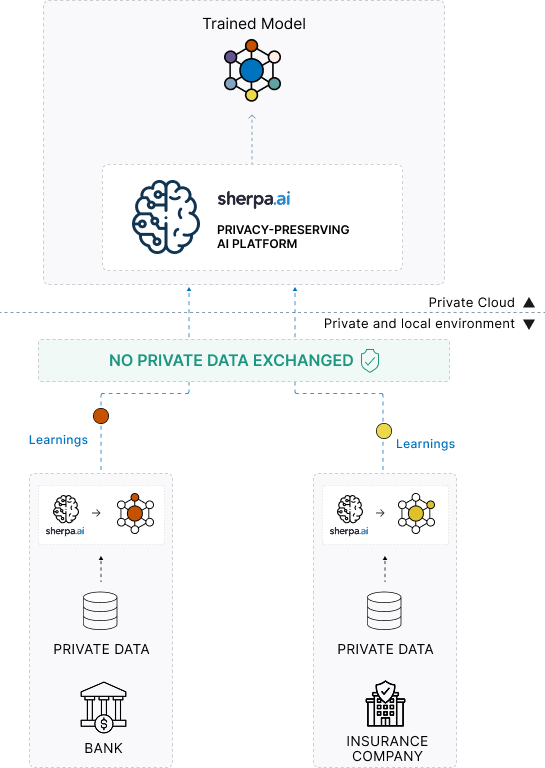}
    \caption{Proposed Architecture for FL.}
    \label{fig:architecture}
\end{figure}

This section describes the federated setup for VFL and SBVFL experiments, detailing the data split across the two nodes and the techniques employed. Below, we propose an architecture for FL designed for collaborative model training between two nodes: a bank and an insurance company (see Figure~\ref{fig:architecture}).

\begin{itemize}
    \item \textbf{Active Node (Bank)}: As in remark \ref{remark_active_party}, the bank is a physical \textit{active node} $\mathcal{F}_{\mbox{\tiny{A}}}$ hosting two logical nodes (the passive node $\mathcal{F}_2$ and the server $\mathscr{S}$), since it possesses both input features and labels.
    The features contained by the Bank include: \verb!BILL_AMT6!, \verb!BILL_AMT5!, \verb!AGE!, \verb!SEX!, \verb!PAY_AMT5!, \verb!PAY_AMT3!, \verb!PAY_5!, \verb!PAY_6!, \verb!PAY_4!, \verb!PAY_AMT6! and the target \verb!default payment next month!;
    \item \textbf{Passive Node (Insurance Company)}: The insurance plays the role of passive node $\mathcal{F}_1$.
    The features contained by the Insurance Company include: \verb!PAY_0!, \verb!BILL_AMT1!, \verb!PAY_AMT2!, \verb!PAY_AMT1!, \verb!PAY_2!, \verb!PAY_AMT4!, \verb!BILL_AMT4!, \verb!BILL_AMT3!, \verb!LIMIT_BAL!, \verb!MARRIAGE!, \verb!BILL_AMT2!, \verb!PAY_3!, \verb!EDUCATION!.
\end{itemize}

\subsection{Experiment} \label{Experiment}

In this section, we detail the models and evaluation methods employed in the experiment, which investigates collaborative training between a bank and an insurance company under various data-sharing scenarios.

We optimized model architectures for each training scenario to maximize predictive performance while adhering to privacy constraints. The configurations are as follows:

\begin{itemize}
    \item For the local (single node) and centralized scenarios, we used a stacked ensemble model combining a Random Forest classifier, a Gradient Boosting classifier, and a Neural Network with an adaptive learning rate to leverage diverse learning strengths.
    \item For the VFL scenario, we employed Neural Networks, as they meet the differentiability requirements of the VFL architecture (see Subsection \ref{subsection_standard_VFL}).
    \item For the SBVFL scenario, we utilized the same stacked ensemble model as in the local and centralized cases, exploiting SBVFL’s flexibility. Additionally, a Neural Network was implemented to enable direct comparison with VFL under identical conditions.
\end{itemize}

Note that VFL’s federated scheme (described in Subsection \ref{subsection_standard_VFL}) mandates models to be differentiable with respect to their parameters, limiting the choice of models compared to SBVFL’s more flexible framework.

\paragraph{Experimentation Setup.}
In all experiments, 20\% of the dataset is reserved as a hold-out test set, remaining unseen during training to ensure unbiased evaluation. Of the training dataset, 75\% comprises shared customer records common to both the bank and the insurance company, representing the intersection of their datasets.

\paragraph{Metrics.}
For evaluation, we adopted the area ratio metric~\cite{YEH20092473} to assess model performance across scenarios. To provide further insight, we also report the AUC and F1-score, which are widely employed in similar classification problems.

\subsection{Results}

This section displays the results of completing several experiments as described in Section \ref{Experiment}

\textbf{Experiment I} 

In this experiment, a single node, centralized, and SBVFL employ the same ML model, a stacked ensemble model combining a Random Forest classifier, a Gradient Boosting classifier, and a Neural Network with an adaptive learning rate to leverage diverse learning strengths.

For standard VFL, we had to adopt a Neural Network, since other models, like the one selected for this experiment, are not supported\footnote{In standard VFL, we have the constraint of differentiability of the ML model with respect to its parameters}.

\begin{figure}[h]
	\centering
        \includegraphics[width=1.0\columnwidth]{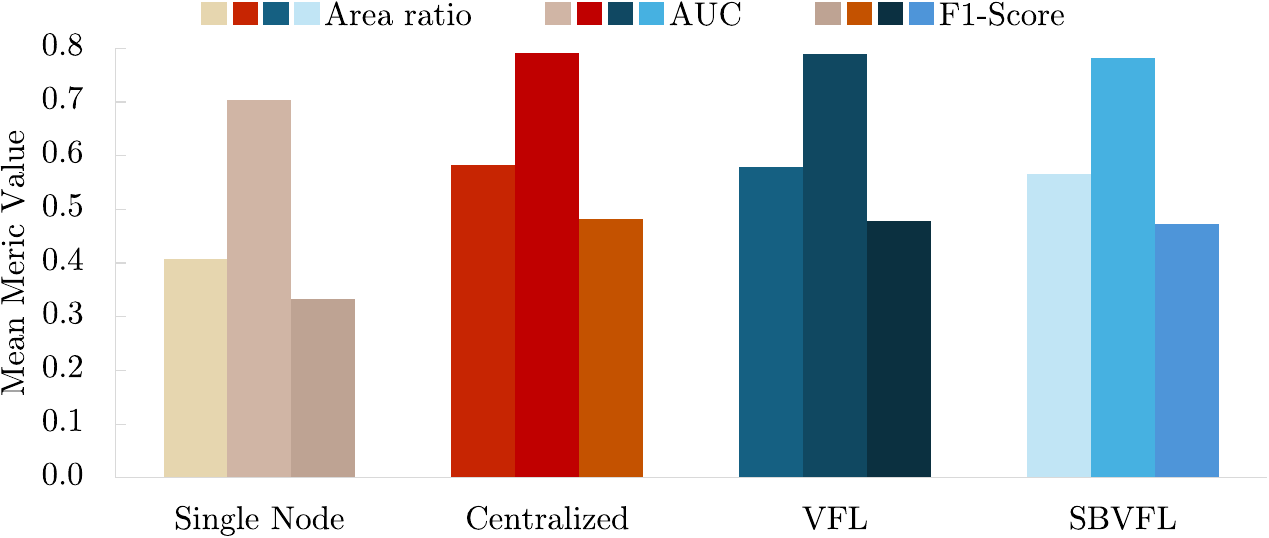}
	\caption{Mean across twenty random trials of single node ML, centralized ML, VFL, and SBVFL. }\label{fig:area_ratio_1002}
\end{figure}

Figure \ref{fig:area_ratio_1002} and Table \ref{tab:table_accuracy2} reveal the performance trade-offs of different models trained with single node, VFL, SBVFL, and Centralized training scenarios using the Sherpa.ai platform. Notice that the centralized case is displayed for benchmarking purposes. However, it is not possible in practice, since it violates privacy regulations.

Table \ref{tab:table_accuracy2} shows that SBVFL outperforms the single node baseline and approaches the performance of VFL and centralized models, while requiring dramatically fewer communications. On the one hand, the centralized scenario is displayed for benchmarking purposes. However, it is not possible in practice since it violates privacy regulations. On the other hand, VFL requires excessive communication, making it unfeasible in practice. SBVFL is then a real-world-ready trade-off between privacy and prediction performances. Indeed, SBVFL requires significantly less communication (6 vs. 6,756 for VFL), proving it as a more efficient methodology. 

%
%
%
%
%
%



\begin{table}[h]
  \centering
  \begin{tabularx}{\columnwidth}{L|Y|Y|Y|Y}
    \toprule
    \textbf{Metric} & \textbf{Single node (STD)} & \textbf{Centralized (STD)} & \textbf{VFL (STD)} & \textbf{SBVFL (STD)} \\
    \midrule
    \textbf{Area ratio} & 0.407 (0.0020) & 0.583 (0.0018) & 0.580 (0.0015) & 0.579 (0.0043)  \\
    \midrule
    \textbf{AUC} & 0.703 (0.0009) & 0.792 (0.0009) & 0.790 (0.0008) & 0.790 (0.0021) \\
    \midrule
    \textbf{F1-Score} & 0.334 (0.0041) & 0.482 (0.0023) & 0.478 (0.0007) & 0.476 (0.0093) \\
    \midrule
    \textbf{Number of communications} & N/A & N/A & 6,756 & 6 \\
    \bottomrule
  \end{tabularx}
  \vspace{10pt}
  \caption{Mean and standard deviation (STD) across 20 experiments of Area Ratio, AUC, and F1-score of local ML (Single Node), Centralized ML, VFL, and SBVFL. The random baseline for the area ratio is $0$, whereas for AUC is $0.5$.}
\label{tab:table_accuracy2}
\end{table}

\textbf{Experiment II} 

The purpose of this experiment is to offer another comparison between VFL and SBVFL. This experiment is conducted to complement Experiment I and facilitate a comparison between the two FL paradigms under the same architecture, but with the model replaced by a Neural Network (ResNet). 

\begin{figure}[h]
    \centering
\includegraphics[width=1.0\columnwidth]{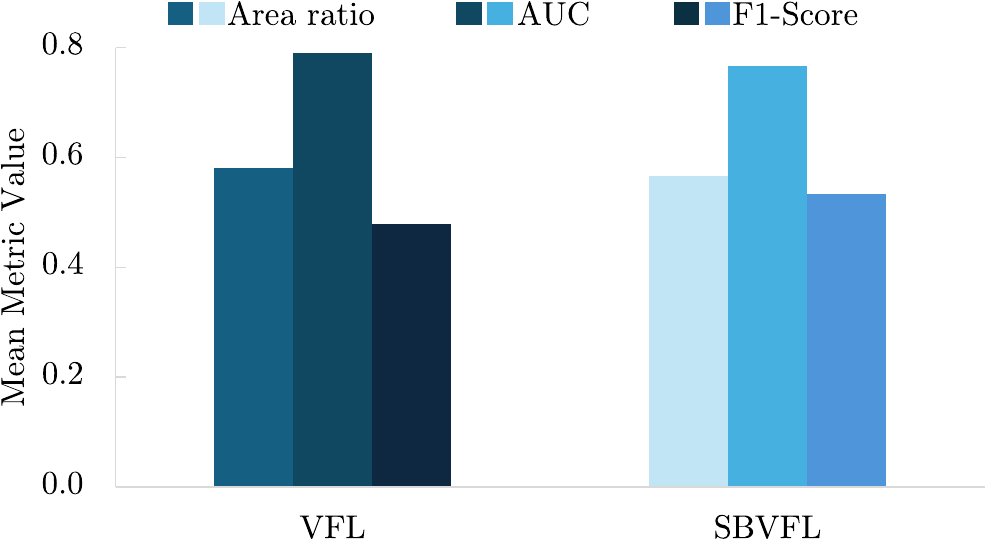}
    \caption{Mean across twenty random trials of VFL and SBVFL.}
    \label{fig:Exp_2}
\end{figure}

The average performance metrics across twenty random trials of the trained model with VFL and SBVFL under the same architecture are displayed in Figure \ref{fig:Exp_2} and Table \ref{tab:Exp2}. SBVFL requires significantly fewer communications, proving to be a more efficient technique.


\begin{table}[h]
  \centering
  \begin{tabularx}{\columnwidth}{L|Y|Y}
    \toprule
    \textbf{Metric} & \textbf{VFL (STD)} & \textbf{SBVFL (STD)} \\
    \midrule
    \textbf{Area Ratio} & 0.580 (0.0015) & 0.532 (0.0116)  \\
    \midrule
    \textbf{AUC} & 0.790 (0.0008) & 0.766 (0.0058)  \\
    \midrule
    \textbf{F1 Score} & 0.478 (0.0007) & 0.534 (0.0079)  \\
    \midrule
    \textbf{Number of communications} & 6,756 & 6 \\
    \bottomrule
  \end{tabularx}
  \vspace{10pt}
  \caption{The mean across twenty random trials of the performances of the trained model under VFL and SBVFL.}
\label{tab:Exp2}
\end{table}

Comparing the SBVFL results in this section with those in Section~\ref{sec:experimental_results} (EMNIST Digits), we find that in this real-world setting the advantages of SBVFL are even more pronounced. In Section~\ref{sec:experimental_results}, the large number of nodes and the higher-dimensional data made federated training more challenging. By contrast, here we:

\begin{itemize}
    \item Demonstrate SBVFL's flexibility across a broad range of ML models.
    \item Observe a substantial reduction in communication rounds, which is critical in real-life distributed deployments.
\end{itemize}

\section{Limitations}\label{sec:Limitations}
One limitation of applying SBVFL is a slight loss in predictive performance (see Sections~\ref{sec:experimental_results} and~\ref{sec:Real world use cases and benefits}). This may be due to the artificially separated manifolds induced by the matrix-valued operator $\Lambda_k$ defined in Equation~\ref{eq:Lambda_k_def}. Specifically, if $Q$ denotes the privacy multiplier, there are $Q-1$ such manifolds. Another limitation arises from the choice of $Q$: while higher values of $Q$ strengthen privacy guarantees, they may also slow convergence or increase variance in training dynamics. This trade-off underscores the need for a balance between preserving privacy and enhancing learning efficiency.

\section{Conclusion}\label{sec:Conclusion}

In this paper, we presented SBVFL, a novel vertical FL method that outperforms existing approaches. The term ``Blind” reflects that clients train locally without continuous coordination with the server. This is enabled by server-issued synthetic labels, privacy-preserving proxy targets constructed so that clients cannot infer the true labels. During server-side aggregation, these proxies are mapped back to the original labels, preserving utility while substantially strengthening privacy and security.

Our experimental results in Section~\ref{sec:experimental_results}, grounded on the solid theoretical foundations of Theorem~\ref{th_BVFL}, demonstrate the effectiveness of SBVFL. The method achieves accuracy levels comparable to both centralized and standard VFL, while dramatically reducing node–server communications during training by $\sim$99\%. SBVFL therefore represents a significant leap forward in preserving security and privacy for distributed heterogeneous training, as discussed in Section~\ref{sec:privacy}. Moreover, this approach unlocks new use cases, as the large number of communications required by the standard VFL approach makes it computationally infeasible.





\section*{Contributions and Acknowledgments} \label{app:A}

Alex Acero

Daniel M. Jimenez-Gutierrez

Dario Pighin

Enrique Zuazua

Joaquin Del Rio

Xabi Uribe-Etxebarria

\vspace{5mm}
The authors are presented in alphabetical order by first name. 

\bibliographystyle{IEEEtran}
\bibliography{my_references_Sherpa.bib}

\end{document}